\RequirePackage{fix-cm}
\documentclass[smallcondensed]{svjour3}     
\smartqed  
\usepackage{graphicx}
\usepackage{amsfonts}
\usepackage{color}
\usepackage{soul}
\usepackage[ruled,linesnumbered]{algorithm2e}
\usepackage{amsmath}
\usepackage[]{subcaption}
\usepackage{blkarray}
\usepackage{xcolor}
\usepackage{url}
\SetKwComment{Comment}{$\triangleright$\ }{}
\SetKwInOut{Parameter}{Parameters}
\DeclareMathOperator*{\argmax}{arg\,max}

\SetKwComment{Comment}{$\triangleright$\ }{}
\DeclareMathOperator*{\hadamand}{\oplus}
\usepackage[numbers,sort&compress]{natbib}
\begin{document}

\title{Propositionalization and Embeddings: \\
Two Sides of the Same Coin
}


\author{Nada Lavra\v{c}         \and
Bla\v{z} \v{S}krlj  \and \\ Marko Robnik-\v{S}ikonja
}


\institute{N. Lavra\v{c} \at
              Jo\v{z}ef Stefan Institute, Jamova 39, 1000 Ljubljana, Slovenia \\
              University of Nova Gorica, Glavni trg 8, 5271 Vipava, Slovenia \\
              \email{nada.lavrac@ijs.si}           
           \and
         B. \v{S}krlj \at
              International Postgraduate School Jo\v{z}ef Stefan, and Jo\v{z}ef Stefan Institute,\\ Jamova 39, 1000 Ljubljana, Slovenia \\  
                         \email{blaz.skrlj@ijs.si}   
                      \and
           M. Robnik-\v{S}ikonja \at University of Ljubljana, Faculty of Computer and Information Science,\\ Ve\v{c}na pot 113, 1000 Ljubljana, Slovenia \\
        \email{marko.robnik@fri.uni-lj.si} 
}

\date{Received: date / Accepted: date}

\maketitle
\setcounter{tocdepth}{2}

\begin{abstract}
Data preprocessing is an important component of  machine learning pipelines, which requires ample time and resources. An integral part of  preprocessing is data transformation into the format required by a given learning algorithm. This paper outlines some of the modern data processing techniques used in relational learning that enable data fusion from different input data types and formats into a single table data representation, focusing on the propositionalization and embedding data transformation approaches. While both approaches aim at transforming data into tabular data format, they use different terminology and task definitions, are perceived to address different goals, and are used in different contexts. This paper contributes a unifying framework that allows for improved understanding of these two data transformation techniques by presenting their unified definitions, and by explaining the similarities and differences between the two approaches as variants of a unified complex data transformation task. In addition to the unifying framework, the novelty of this paper is a unifying methodology combining propositionalization and embeddings, which benefits from the advantages of both in solving complex data transformation and learning tasks. We present two efficient implementations of the unifying methodology: an instance-based PropDRM approach, and a feature-based PropStar approach to data transformation and learning, together with their empirical evaluation on several relational problems. The results show that the new algorithms can outperform existing relational learners and can solve much larger problems.
\end{abstract}

\section{Introduction}

Data preprocessing for machine learning is a great challenge for a data scientist faced with large quantities of data in different forms and sizes. Most of the modern data processing techniques enable data fusion from different data types and formats into a single table data representation, which is expected by standard machine learning techniques including rule learning, decision tree learning, support vector machines (SVMs), deep neural networks (DNNs), etc. The key element of the success of modern data transformation methods is that similarities of  original instances and their relations are encoded as distances in the target vector space.

Two of the most prominent data transformation approaches outlined in this paper are propositionalization and embeddings.
 While propositionalization \citep{Propositionalization,zelezny2006} is a well known data transformation technique used in relational learning (RL) and inductive logic programming (ILP)  \cite{ILP-book,ILP-Lavrac,deraedtbook:2008}, embeddings \cite{word2vec,Wu2018} have only recently been recognized by RL and ILP researchers as a powerful technique for preprocessing relational and complex structured data. In the relational learning context of this paper, both approaches take as input a relational data set (e.g., a given relational database) and transform it into a single data table format, which is then used as an input to a propositional learning algorithm of choice.

The first aim of this paper is to present a unifying survey of propositionalization and embedding data transformation approaches. While both approaches aim at transforming data into a tabular data format, the approaches use different terminology and task definitions, claim to have different goals, and are used in very different contexts.
This paper contributes an improved understanding of these data transformation techniques by presenting a unified terminology and definitions, by explaining the similarities and differences of the two definitions  as variants of a unified complex data transformation task, by exploring the apparent differences between the two approaches, and by outlining some of their advantages and disadvantages. 

In addition to the unifying survey, the main novelty of this paper is a unifying methodology that combines propositionalization and embeddings, which benefits from the advantages of both in solving complex data transformation and learning tasks. The unifying methodology resulted in two new pipelines, PropDRM  and  PropStar, which implement an instance-based and a feature-based approach to data transformation and learning, respectively. Both approaches are computationally efficient and can successfully solve much larger tasks than the existing relational learning approaches. We made their code publicly available.

The paper starts by motivating the need for transforming heterogeneous relational data into a tabular format in Section~\ref{sec-motivation}.
Section~\ref{sec-transf-represent}
introduces the data transformation approaches in the context of information representation levels proposed by  G\"{a}rdenfors \cite{Gardenfors:2000}. Section~\ref{sec:related} presents the related work, 
focusing on selected propositionalization and embeddings methods relevant to the relational learning context of this paper.
Section~\ref{sec-unifyingFramework} presents a unifying framework for propositionalization and embeddings, allowing for the analysis of characteristic properties of these data transformation approaches. Section~\ref{sec:implementations} proposes a unifying methodology that combines propositionalization and embeddings, which benefits from the advantages of both, and presents two implementations of the proposed unifying framework: an instance-based embedding approach PropDRM based on the existing Deep Relational Machines (DRM) \citep{srinivasan209jmlr,lodhi2013deep}, followed by a novel feature-based embedding approach PropStar proposed in this paper, using the StarSpace entity embedding approach \citep{Wu2018}.
Experimental evaluation of the proposed implementations is presented in Section~\ref{sec:experimentalEvaluation}.
The paper concludes by a summary and some ideas for future work in Section~\ref{sec-conclusions}.


\section{Motivation}
\label{sec-motivation}

Machine learning is the key enabler for computer systems to progressively improve their performance when helping humans to solve difficult problem solving tasks. Nevertheless, current machine learning approaches only come half-way in helping humans, as humans still have to formulate the problem and prepare the data in the form that is best suited to the powerful machine learning algorithms. 

Most of the best performing machine learning algorithms, like Support Vector Machines (SVMs) or deep neural networks, assume numeric data and 
outperform symbolic approaches in terms of predictive performance, efficiency, and scalability. The dominance of numeric algorithms started in 1980s with the advent of backpropagation and neural networks \cite{Rumelhart1986}, continued in late 1990s and early 2000s with SVMs \cite{cortes95}, 
and finally reached the current peak with deep neural networks \cite{Goodfellow-et-al-2016}. Deep neural networks are currently considered the most powerful learners for solving many of previously unsolvable learning problems in computer vision (face recognition rivals humans' performance), game playing (a program has beaten a human champion in the game of Go), and natural language processing (successful automatic speech recognition and machine translation).

While the most powerful machine learning approaches are  numeric, humans perceive and describe  real-world problems mostly in symbolic terms, using various data representation format, such as graphs, relations, texts or electronic health records,
all involving discrete representations.
However, if we are to harness the power of successful numeric deep learning approaches for discrete learning problems, 
discrete data should be transformed into a form suitable for numeric learning algorithms. 
The viewpoint of addressing real-world problems as numeric has a rationale even for discrete domains, as many symbolic learners perform generalizations based on object similarity. For example, in graphs, nodes can represent similar entities or have connections with similar other nodes; in text, words can appear with similar contexts or play the same role in sentences; in medicine, patients may have similar symptoms or similar disease histories. Such similarities are used by numerous machine learning algorithms to generalize and learn, including classical bottom-up learning approaches such as hierarchical clustering, as well as symbolic learners adapted to top-down induction of clustering trees \citep{Blockeel98top-downinduction}. If we want to exploit the power of modern machine learning algorithms, like SVMs and deep neural networks, to process the inherently discrete data, one has to transform discrete data into (numeric) vectors in such a way that similarities between objects are preserved and encoded as distances in the transformed (numeric) space.

Contemporary preprocessing approaches that prepare numeric vector data for machine learning algorithms are called \emph{embeddings}. Nevertheless, as demonstrated in this paper, symbolic data transformations, as ancestors of the contemporary embedding approaches, remain relevant: the role of \emph{propositionalization}, a symbolic approach to relational data transformation into feature vectors, is not only to enable contemporary machine learning algorithms to induce better predictive models, but to allow descriptive data mining approaches to discover interesting  human-comprehensible patterns in symbolic data.

As this paper demonstrates, albeit propositionalization and embeddings represent different types of data transformations, these approaches actually represent the \emph{two sides of the same coin}. The main unifying element they have in common is that they transform the data into a vector format and encode the relations between objects in the original space as distances in the new vector space.

\section{Data transformations and information representation levels}
\label{sec-transf-represent}

As this section will show, we consider data transformations as a particular subprocess of data preprocessing. Data preprocessing aims to handle missing attribute values, control out-of-range values and impossible attribute-value combinations, or handle noisy or unreliable data, to name just some of the types of data irregularities addressed in processing real-life data. Data preprocessing may include data cleaning, instance selection, normalization, feature engineering (feature extraction and/or feature construction), data transformation, feature selection, etc. The result of data preprocessing is the final training set, which is  used as input to a machine learning algorithm. 

Data preprocessing can be manual, automated, or semi-automated. We focus on automated transformations of data, present in heterogeneous types and formats, into a uniform tabular data representation. We refer to this specific automated data preprocessing task as \emph{data transformation}, and define it as follows.

\begin{definition}[Data transformation]
\label{def-transformation}

\emph{Data transformation} is a step in the data preprocessing task that 
\index{data transformation|(}%
automatically transforms the input data and the background knowledge into a uniform tabular representation, where each row represents a data instance, and each column represents one of the dimensions in a multi-dimensional feature space.

\end{definition}

In the above definition, we decided to distinguish between \emph{data} and \emph{background knowledge}. This is an intentional decision, although it could be argued that in some settings, we could refer to both as data. Let us provide an operational distinction between data and background knowledge. \emph{Data} is considered by the learner as the target data from which the learner should learn a model (e.g., a classifier in the case of class labeled data) or a set of descriptive patterns (e.g., a set of association rules in the case of unlabeled data). \emph{Background knowledge} is any additional knowledge used by the learner in model or pattern construction from the target data. Simplest forms of background knowledge define hierarchies of features (attribute values), such as 
color \emph{green} being more general than \emph{light green} or \emph{dark green}. More complex background knowledge refers to any other declarative prior domain knowledge, such as knowledge encoded in relational databases, knowledge graphs or domain specific taxonomies and ontologies, such as the Gene Ontology, in its 2020-05-02 release including 44,508 GO terms, 7,765,270 annotations, 1,464,358 gene products and 4,593 species.

This data transformation setting is applicable in various data science scenarios involving relational data mining, inductive logic programming, text mining, graph and network mining as well as tasks that require fusion of data of a variety of data types and formats and their transformation into a joint data representation formalism. 

\subsection{Information representation levels}
\label{sec-Gardenfors}

As currently the most powerful machine learning (ML) algorithms take as input numeric  representations, users of ML algorithms tend to transform other forms of human knowledge into the numeric representation space. Interestingly, even if this is countering a standard RL and ILP viewpoint, this is true also for symbolic representations, which are currently used to store most of the human knowledge. 

The distinction between the symbolic and numeric representation space mentioned above can be further clarified in terms of 
the \emph{levels of cognitive representations}, introduced by \citet{Gardenfors:2000}, i.e. the neural, spatial and symbolic representation levels. In his theory, G\"{a}rdenfors assumes that when modeling cognitive systems in terms of information processing, all three levels are connected: starting from the sensory inputs at the lowest neural representation level, resulting in spatial representations at the middle conceptual spaces level, up to symbolic representations at the level of language. 
\begin{description}
\item[\emph{Neural}.] This representation level corresponds to the sub-conceptual connectionist level. At this level, information is represented by activation patterns in densely connected networks of primitive units. This enables concepts to be learned from the observed data by modifying the connection weights between the units.
\item[\emph{Spatial}.] This representation level is modeled in terms of G\"{a}rdenfors' conceptual spaces. At this level,  information is represented by points or regions in a conceptual space built upon some dimensions that represent geometrical, topological or ordinal properties of the observed objects. In spatial representations, the similarity between concepts is represented in terms of the distances between the points or regions in a multidimensional space, where concepts are learned by modeling the similarity between the observed objects.
\item[\emph{Symbolic.}] At this representation level, information is represented by the language of symbols (words), where the meaning is internal to the representation itself (i.e. symbols have meaning only in terms of other symbols, while their semantics is grounded in the spatial level), and concepts are learned by symbolic generalization rules. 
\end{description}



From the perspective of this paper, the above levels of cognitive representations introduced by \citet{Gardenfors:2000} provide a theoretical ground to separate the learning approaches as well as the data transformation approaches into three categories based on the levels of their output representation space: neural, spacial and symbolic. However, given the scope of this paper, we do not consider \emph{neural transformations}, and focus only on two data transformation types:
\begin{itemize}
\item \emph{symbolic transformations}, in this paper referred to as \emph{propositionalization}, denoting data transformations into a symbolic representation space,  and
\item \emph{numeric transformations}, in this paper referred to as \emph{embeddings}, denoting data transformations into a spatial representation space.
\end{itemize}

These two data transformation approaches are briefly introduced below, and further described in the related work (Section~\ref{sec:related}).

\subsection{Transformations into symbolic representation space}
\label{sec-symbolicTransform}



The past decades of machine learning were characterized by symbolic learning, where the result of a machine learning or data mining algorithm was a predictive model of a set of patterns described in a symbolic representation language, resulting in symbolic human-understandable patterns and models. Symbolic machine learning approaches include rule learning \cite{AQ15,CN2},  decision tree learning
  \cite{C4} and learning logical representations by relational learning and inductive logic programming (ILP) algorithms \cite{ILP-book,ILP-Lavrac,deraedtbook:2008}. 

To be able to apply a symbolic learner, the data is typically  transformed into a single tabular data format, where each row represents a single data instance, and each column represents an attribute or a feature. Such transformation into symbolic vector space (i.e. a symbolic data table format) is well known in the ILP and relational learning community, where it is referred to as \emph{propositionalization}. Propositionalization approaches are presented in Section~\ref{sec:propositionalization}.


%
%

\subsection{Transformations into numeric representation space}
\label{sec-numericTransform}

In the last 20 years we have been witnessing increasing dominance of statistical machine learning and 
pattern-recognition methods, including 
\index{neural networks}neural network
  learning \cite{PDP}, \index{support vector machine}Support Vector Machines (SVMs)
  \cite{StatisticalLearningTheory-Nature,LearningWithKernels}, 
  random forests \cite{RandomForests}, and boosting \cite{AdaBoost}.
 These statistical approaches are quite different from the symbolic approaches mentioned in Section~\ref{sec-symbolicTransform}, however there are many approaches that cross these
boundaries, including e.g., the CART decision
tree learning algorithm \cite{CART}, the Bump hunting rule
learning algorithm \cite{BumpHunting}, which are
firmly based in \index{statistics}statistics. Moreover, ensemble techniques such as
\index{boosting}boosting \cite{AdaBoost}, \index{bagging}bagging \cite{Bagging} or \index{random forests}random forests
\cite{RandomForests} also combine the predictions of multiple logical
models on a sound statistical basis
\cite{Boosting-Margin,Boosting-Statistical1,Boosting-Statistical2}. 
All these are also considered to belong to the family of statistical learning approaches.

To be able to apply a statistical learner, the data is typically transformed into a single tabular data format, where each row represents a single data instance, and each column is a numeric attribute or a numeric feature, with some predefined range of numeric values. Such transformation into numeric vector space (i.e. a numeric data table format) is well known in the deep learning community, where it is referred to as \emph{embedding}. Approaches to embedding relational structures are presented in Section~\ref{sec:emb-rel}.


%

\section{Related work}
\label{sec:related}


In this section we first outline various transformation methods in Section~\ref{sec-transformationMethods}, followed by a more detailed description of the data transformation methods relevant for the context of relational learning, i.e. propositionalization and embeddings, in Section~\ref{sec:propositionalization} and Section~\ref{sec:emb-rel}, respectively.

\subsection{Outline of data transformation methods}
\label{sec-transformationMethods}

While there are many algorithms for transforming data into a spatial representation, it is interesting that recent approaches rely on deep neural networks, thereby harnessing the neural representation level as the means to transform symbolic representations into the spatial representation.  Below we list the main types of approaches that perform transformations between representations. 

\begin{description}
\item[Community detection and graph traversal methods.] Many complex data sets can be represented as graphs, where nodes represent data instances and edges represent their relations. Graphs can be homogeneous (consisting of a single type of nodes and relations) or heterogeneous (consisting of different types of nodes and relations). To encode a graph in a tabular form by preserving the information about the relations, various graph encoding techniques were developed, such as propositionalization via random walk graph traversal, representing nodes via their neighborhoods and communities \cite{community}. These approaches are frequently used for data fusion in mining heterogeneous information networks. Neural network approaches (presented below) are also very competitive as means for encoding graphs.

\item[Matrix factorization methods.] When data is not  explicitly presented in the form of relations but the relations between objects are implicit, given by a similarity matrix, the objects can be encoded in a numeric form using matrix factorization. As an example take Latent Semantic Analysis used in text mining, which factorizes a word similarity matrix to represent words in a vector form. Another example is factorization of graph adjacency matrices. These types of embeddings were largely superseded by deep neural networks which, instead of observing similarity between different objects, construct a prediction task and forecast similarity.  For example, for text, given a word, the word2vec  embedding method \cite{word2vec} predicts words in its neighborhood.  

\item [Propositionalization methods]  are used to get tabular data from multirelational databases as well as from a mixture of tabular data and background knowledge in the form of logic programs or networked data, including ontologies. These transformations were mostly developed within the Inductive Logic Programming and Relational Learning community, and are still actively researched and used. Propositionalisation methods do not perform dimensionality reduction and are most often used with data mining and symbolic machine learning algorithms. We discuss these methods in Section \ref{sec:propositionalization}.

\item[Neural networks based methods.]
In neural networks the information is represented by activation patterns in interconnected networks of primitive units. This enables that concepts are gradually learned from the observed data by modifying the connection weights between the hierarchically organized units. These weights can be extracted from neural networks and used as a spatial representation that transforms relations between entities into distances. Recently, this approach became a prevalent way to build representation for many different types of entities, e.g., texts, graphs, electronic health records, images, relations, recommendations, etc. 
In Section~\ref{sec:emb-rel} we describe the data types and approaches, which are capable of embedding relational structures and are therefore most relevant for the context of this paper. These include knowledge graph embeddings (presented in Section~\ref{sec:kge}), entity embeddings capable of forming (both supervised and unsupervised) representations based on the similarity of entities (presented in  Section~\ref{sec:entity}), and Deep Relational Machines methodology that links symbolic representations to deep neural networks (presented in Section~\ref{sec:drm}).

\item[Other embedding methods.] Other forms of  embeddings were developed by different communities that observed the need to better represent the (symbolic) data. 
For example, Latent Dirichlet Allocation (LDA) \cite{blei2003latent} used in text analysis learns distributions of words for different topics. These distributions can be used as an effective embedding for words, topics, and documents. Feature extraction methods form a rich representation of instances by projecting them into a high dimensional space \cite{Lewis:SIGIR-92}.
Another example of (implicit) transformation into high dimensional space is the kernel convolutional approach proposed by \citet{haussler1999convolution}, which introduces the idea that kernels can be used for discrete structures by iteratively applying convolution and kernels to smaller parts of the data structure. Convolutional kernels exist for sets, graphs, trees, strings, logical interpretations, and relations \citep{cumby2003kernel}. This allows methods such as SVM or Gaussian Processes to work with  relational data.
 Most of these embeddings are recently superseded or merged with neural networks. 

\end{description}

All the above approaches perform data transformations from different data formats to a single table representation. However, their underlying principles are different: while  factorization and neural embeddings perform dimensionality reduction, resulting in  lower-dimensional feature vector representations capturing the semantics of the data, propositionalization results in a vector representation using relational features with a higher generalization potential than the features used in the original data representation. Note that there exist also other approaches to data transformation and fusion, including HINMINE \cite{Kralj2018}, metapath2vec \cite{zhu2018prediction} and OhmNet \cite{zitnik2017predicting}, which are out of the main scope of this paper. 

\subsection{Propositionalization}
\label{sec:propositionalization}

In propositionalization, relational feature construction is the most common approach to data transformation. 
LINUS~\citep{Linus} was one of the pioneering propositionalization approaches using automated relational feature construction. 
LINUS was restricted to generation of features that do not allow recursion and existential local variables, which means that the target relation cannot be many-to-many and self-referencing.  The second limitation was more serious:
the queries could not contain joins (conjunctions of literals).  The LINUS descendant SINUS \citep{Linus-Extended} incorporates more advanced feature
construction techniques inspired by 
1BC~\citep{flach19991bc}. The LINUS approach had many followers, including relational subgroup discovery system RSD~\citep{zelezny2006}, which is outlined also in the list of propositionalization approaches below.
Alternatives to  relational feature construction include the construction of aggregation queries. 

In this section we first clearly define the distinction between attributes and features, followed by an outline of selected propositionalization approaches and of the specific Wordification approach used in the algorithms developed in this work.

\subsubsection{Features}
\label{sec-features}

To be able to apply a symbolic propositional learner, the data should be represented in a single table data format, where each row represents a single data instance, and each column represents an attribute or a feature. For the sake of clarity, let us distinguish between \emph{attributes} and \emph{features} below.

Attributes
that describe the data instances can be either  numeric
variables (with values like $7$ or $1.5$) or nominal/discrete
variables (with values like \emph{red} or \emph{female}).
In contrast to attributes, a \emph{feature} describes the presence or absence of some property of an instance.
As a result, features are always
Boolean-valued (values \emph{true} or \emph{false}).
For example, for attribute \emph{gender} with values \emph{female} and \emph{male}, two separate features can be constructed: $f_1$: \emph{gender}=\emph{female} and $f_2$: \emph{gender}=\emph{male}, and only one of these features is assumed to be \emph{true} for an individual data instance.
Note that features
are different even from  binary-valued attributes:
e.g.,
for a binary attribute $a_{i}$ with values \emph{true} and \emph{false}, there are two
corresponding features: $f_3$: $a_{i}=$ \emph{true} and
$f_4$: $a_{i}=$~\emph{false}.
Furthermore, features can test a
value of a single attribute, like $a_{j}>3$, or
they can represent complex logical and numerical relations,
integrating properties of multiple attributes, like $f_5$: $a_{k}< 2\cdot(a_{j}-a_{i})$.

Previous feature types are referred to as
\emph{propositional features}. On the other hand, \emph{relational features} relate the values of different attributes to
each other. In the simplest case, for example, they test for the equality or
inequality of the values of two attributes of the same type, such as \emph{Length} and \emph{Height}. More complex relational features can use the background relations, e.g.,  $f_6$: \emph{adjacent(NodeX, NodeY)}. Even more advanced, relational features can introduce new variables. For example, if relations are
used to encode a graph, a relational feature  such as $f_7$: 
\emph{color(CurrentNode, blue)} $\wedge$ \emph{link(CurrentNode, NewNode)} $\wedge$ \emph{color(NewNode, red)}, 
can introduce a new variable \emph{NewNode} to subsequently test whether there exists a previously not visited node in the graph that is colored red.

Take a simple toy trains example learning problem illustrated in Appendix~\ref{sec-wordificationExample}, and two complex relational features describing trains: \\ \\
\indent $f_8$: hasCar(T,C) $\wedge$ carLength(C,short) $\wedge$ carRoof(C,peaked) \\
\indent  $f_9$: hasCar(T,C1) $\wedge$ carLength(C1,short) $\wedge$ hasCar(T,C2)  $\wedge$ carRoof(C2,peaked) \\

\noindent Feature $f_8$ is a single complex relational feature, while  $f_9$ contains two distinct relational features. Formally, a feature is defined as a minimal set of literals such that it introduces at most one local (i.e. existential) variable in the feature set composing the relational feature. 

The main point of relational features is that they localize variable sharing: 
this can be made explicit by naming the features: \\ \\
\indent  $f_{10}$: hasShortCar(T) $\leftarrow$  hasCar(T,C) $\wedge$ clength(C,short) \\
\indent  $f_{11}$: hasPeakedroofCar(T) $\leftarrow$  hasCar(T,C) $\wedge$ carRoof(C,peaked) \\ 


The propositionalization approach to relational learning captures exactly this idea: generating complex features, such as $f_8$, $f_{10}$ and $f_{11}$, 
which will allow multi-relational data representation of properties of target instances (such as trains $T$) through representations of properties of their components (such as cars $C$). Selected propositionalization approaches, which use complex feature construction in the automated multi-relational data transformation process are outlined below.

\subsubsection{Outline of selected propositionalization algorithms}

Below we outline a selection of propositionalization approaches, while an interested reader can find extensive overviews of different feature construction approaches in the work of \citet{Propositionalization} and \citet{Propositionalization-Comparison}.

\begin{description}

\item[Relaggs~\citep{krogel2001transformation}] stands for \emph{rel}ational \emph{agg}regation. It is a propositionalization approach that takes the input relational database schema as a basis for a declarative bias, using optimization techniques usually used in relational databases (e.g., indexes). The approach employs aggregation functions in order to summarize non-target relations with respect to the individuals in the target table.

\item[1BC~\citep{flach19991bc}] strives to enable the propositional naive Bayes classifier to handle relational data. It does so by a transformation in which a set of first-order conditions is generated and then used as attributes in the naive Bayes classifier.  The transformation, however, is done in a dynamic manner, as opposed to standard propositionalization, which is performed as a static step of data preprocessing. This approach is extended by 1BC2~\citep{DBLP:conf/ilp/LachicheF02}, which allows distributions over sets, tuples, and multisets, thus enabling the naive Bayes classifier to consider also structured individuals.

\item[Tertius~\citep{Tertius}] is a top-down rule discovery system, incorporating first-order clausal logic.  The main idea is that no particular prediction target is specified beforehand, hence Tertius can be seen as an ILP system that learns rules in an unsupervised manner.  Its relevance for this survey lies in the fact that Tertius encompasses 1BC, i.e. relational data is handled through 1BC transformation.

\item[RSD~\citep{zelezny2006}] is a relational subgroup discovery algorithm composed of two main steps: the propositionalization step and the (optional) subgroup discovery step. The output of the propositionalization step can be used also as input to other propositional learners. RSD effectively produces an exhaustive list of first-order features that comply with the user-defined mode constraints, similar to those of Progol~\citep{muggleton1995} and Aleph~\citep{aleph}. Furthermore, RSD features satisfy the connectivity requirement, which imposes that no feature can be decomposed into a conjunction of two or more features.  Mode declarations define the
algorithm's syntactic bias, i.e. the space of possible features.

\item[HiFi~\citep{kuzelka.ilp.2008}] is a propositionalization approach that constructs first-order features with hierarchical structure. Due to this feature property, the algorithm performs the transformation in polynomial time of the maximum feature length. Furthermore, the resulting features are the shortest in their semantic equivalence class. The algorithm is shown to perform several orders of magnitude faster than RSD for higher feature lengths.

\item[RelF~\citep{kuzelka2011}] is the most relevant of the algorithms in the
TreeLiker software~\citep{kuzelka2011}. It constructs a set of tree-like relational features by combining smaller conjunctive blocks. RelF preserves the monotonicity of feature reducibility and redundancy (instead of the typical monotonicity of frequency), which allows the algorithm to scale far better than other state-of-the-art propositionalization algorithms.

\item[Cardinalization~\citep{DBLP:journals/eswa/AhmedLCJB15}] is specifically designed to enable more than just categorical attributes in propositionalization. Specifically, it can handle a threshold on numeric attribute values and a threshold on the number of objects satisfying the condition on the attribute simultaneously.  Cardinalization can be seen as an implicit form of discretization. While in discretization one sets a threshold on a numeric attribute and see how many objects satisfy the threshold later, and the cardinality follows implicitly from the attribute value threshold; on the other hand, in cardinalization, we set a threshold on the cardinality, and let an attribute-value learner decide where the threshold value on the numerical attribute should lie.  Hence, Cardinalization allows for context-aware discretization. Quantiles~\citep{DBLP:journals/eswa/AhmedLCJB15} is a variant of Cardinalization. Instead of choosing an absolute number as cardinality threshold, Quantiles uses a relative number.

\item[CARAF~\citep{2015Charnay}] approaches the problem of large relational feature search space by aggregating base features into complex compounds, which makes CARAF similar to Relaggs.  Complex aggregates run the risk of overfitting.  While Relaggs tackles this problem by restricting itself to relatively simple aggregates, the distinguishing feature of CARAF is that instead it incorporates more complex aggregates into a random forest, which ameliorates the overfitting effect.

\item[Aleph~\citep{aleph}] is the most popular ILP algorithm and is actually an ILP toolkit with many modes of functionality: learning of theories, feature construction, incremental learning, etc.  Aleph uses mode declarations to define the syntactic bias. Input relations are Prolog clauses, defined either extensionally or intensionally. Aleph's feature construction functionality also means it is a propositionalization approach.

\item[Wordification~\citep{perovvsek2013wordification,perovsek2015}] is a propositionalization method inspired by text mining that can be viewed as a transformation of a relational database into a corpus of text documents.  The distinguishing property of Wordification is its efficiency when used on large relational data sets and the potential for using text mining approaches on the transformed propositional data.  While most of the outlined propositionalization algorithms construct complex relational features including variables in the arguments of relational features, Wordification constructs simple, easily interpretable features that are treated as `words' in the transformed \mbox{Bag-Of-Words} representation. It constructs features of the kind $a_i = v_{ij}$ (formulated as  $a_i\_v_{ij}$). In addition to such simple features, it constructs also conjuncts (of size 2) of such features, e.g.,  $a_i = v_{ij}$ $\wedge$ $a_k = v_{kl}$, formulated as $a_i \_v_{ij}\_\_a_k \_ v_{kl}$.  
To avoid confusion in case the same attribute name appeared in several tables, the actual form of features is $t\_a_i\_v_{ij}$ including the indicator of the name of table $t$ in which attribute $a_i$ appears. 
For a simple example of how such features are generated, the reader is referred to Appendix~\ref{sec-wordificationExample}.

\end{description}

\subsubsection{Wordification}
\label{sec-wordificationOutline}
Given that in a previous experimental evaluation of propositionalization algorithms \citep{perovvsek2013wordification,perovsek2015} the Wordification algorithm was shown to be the most effective, we selected Wordification as the propositionalization algorithm of choice in the proposed implementations combining propositionalization and embeddings in Section~\ref{sec:implementations}, where the Wordification algorithm was adapted to handle large data sets. 

In the Wordification implementation, described in detail in Section~\ref{sec:wordification-technical}, the original feature representation \emph{TableName}$\_$\emph{AttributeName}$\_$\emph{AttributeValue} was---for implementational convenience---replaced by a tuple representation (\emph{t.name, c, v}), where \emph{t.name} refers to a table name, $c$ to a given colon (attribute) in the table $t$, and $v$ to a given value $v$ of attribute $c$. Such features will be referred to as \emph{features} or as \emph{relational items} in the algorithm description, as appropriate.

Using this feature representation, Wordification of a multi-relational database can be summarized as the following operation:

\begin{equation*}
\textsc{DB}_i = \biguplus_{t \in \mathcal{T}} \textsc{WORDIFY} (t(m(i)))
\end{equation*}
\noindent where $m$ maps a given table $t$'s indices to target (initial) table indices ($i$) and $\mathcal{T}$ is the set of all tables from which a foreign key path exists to the target table. The $\uplus$ operator represents a disjoint union of multisets (sum), yielding a single multiset (duplicates are allowed).

Foreign keys are designated columns that link data between distinct tables. Value of a foreign key in a given table is referred to as the instance id (the row is uniquely determined by this value). 
Let $C$ represent the set of all columns that are not foreign keys, ids or target classes. The \textsc{WORDIFY} method returns a multiset (a bag) of relational items (for the $i$-th instance) constructed as follows:
\begin{equation*}
\textsc{WORDIFY}(t(m(i))) =  \biguplus_{v \in t[m(i)][c \in C]} (t.name, c, v)
\end{equation*}
\noindent where $t[c]$ represents the values $v$ of table $t$ in column $c$, and $t.\textrm{name}$ is the name of table $t$.
Thus, Wordification is na\"ive in the sense that it simply concatenates attribute values across tables by maintaining the column and table name information in constructing features. The original implementation, however, can become spatially intractable (see \citep{perovvsek2013wordification}, proof of complexity) as its spatial complexity is $\mathcal{O} ( \textrm{row} \cdot \textrm{tables} \cdot 2^{\textrm{col}} )$.
Details of a more efficient implementation of Wordification are available in Section~\ref{sec:wordification-technical}.

\subsection{Embedding relational structures}
\label{sec:emb-rel}

In this section, we discuss methodologies capable of embedding relational structures.
We start with an introduction to knowledge graph embeddings,  an emerging group of methods that operate on large, real-world, annotated graphs, in  Section~\ref{sec:kge}. 
We proceed by the presentation of entity embeddings, a more general methodology capable of supervised, as well as unsupervised embeddings of many entities, including  texts and knowledge graphs in Section~\ref{sec:entity}. Finally, in Section~\ref{sec:drm}, we present Deep Relational Machines, an emerging methodology that links symbolic representations to deep neural networks. 

\subsubsection{Knowledge graph embeddings}
\label{sec:kge}
In knowledge graphs (KG), edges correspond to relations between entities (nodes) and the graphs present Subject-Predicate-Object \emph{triplets}. The KG handling algorithms attempt to solve the problems like triplet completion, relation extraction, and entity resolution. The KG embedding algorithms, briefly discussed below, outline some of the key ideas which render these methods highly scalable and useful for large, semantics-rich graphs.  For detailed description and a recent, extensive overview of the field, we refer the reader to \cite{8047276}, from where we next summarize some of the key ideas underlying knowledge graph embedding.

\begin{figure}[ht]
\centering
\includegraphics[width=0.7\linewidth]{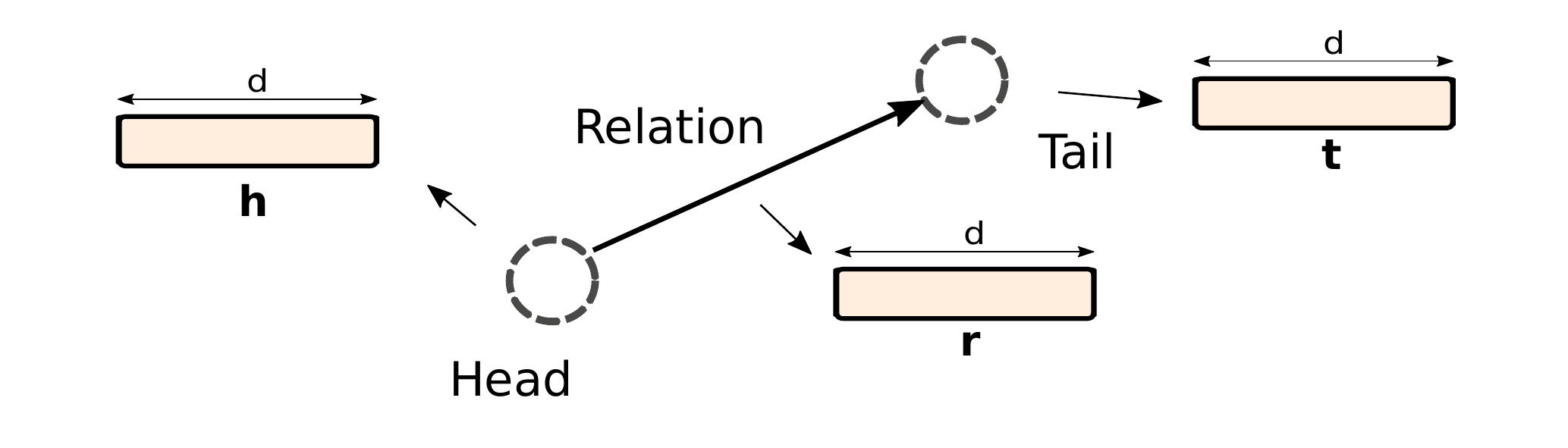}
\caption{Schematic representation of knowledge graph embedding. Head-Relation-Tail $(h,r,t)$ triplets are used as inputs. Triplets are embedded in a common $d$-dimensional vector space.}
\label{fig:kge}
\end{figure}

In the below description of KG embedding algorithms, the Subject-Predicate-Object triplet notation is replaced by the $(h,r,t)$ triplet notation, where $h$ is referred to as the \emph{head} of a triplet, $t$ as the \emph{tail}, and $r$ as the \emph{relation} connecting the head and the tail. A schematic representation of triplet embedding is shown in Figure~\ref{fig:kge}. The embedding methods briefly outlined below optimize the total plausibility of the input set of triplets, where plausibility of a single triplet is denoted with $f_r(h,t)$.

\begin{itemize}
\item The first group of KG embedding algorithms are termed \emph{translational distance models}, as they exploit distance-based scoring functions. They measure the plausibility of a fact as the distance between the two entities, usually after a translation carried out by the relation.   One of the representative methods for this type of embedding is transE \cite{bordes2013translating}, where the cost function being optimized can be stated as:
\begin{equation*}
f_r(h,t) = - \lvert \lvert \textbf{h} + \textbf{r} - \textbf{t} \rvert \rvert^2.
\end{equation*}
For vectors $\textbf{h}$, $\textbf{r}$, and $\textbf{t}$ in the obtained embedding, score $f_r(h,t)$ is high if triplet $(h,r,t)$ is present in the data.
\item The second group of KG embedding algorithms is not deterministic, as it takes into account the uncertainty of observing a given triplet. A representative method for this type of embeddings is KG2E \citep{he2015learning}, which models the triplets with multivariate Gaussians. It models individual entities, as well as relations as vectors, drawn from multivariate Gaussians, assuming that  $\textbf h$, $\textbf r$ and $\textbf t$ vectors are normally distributed, with mean vectors $\mu_h,\mu_r, \mu_t\in\mathbb R^d$ and covariance matrices $\Sigma_h,\Sigma_r,\Sigma_t\in\mathbb R^{d\times d}$, respectively. 
KG2E uses Kullback-Liebler divergence to directly compare the distributions as follows:
\begin{align*}
f_r(h,t) &= \textsc{KL}(\mathcal{N}(\mu_t - \mu_h), \mathcal{N}(\mu_r)) \\ &=\int \mathcal{N}_x (\mu_t - \mu_h, \Sigma_t + \Sigma_h) \ln \frac{\mathcal{N}_x (\mu_t -\mu_h, \Sigma_t + \Sigma_h)}{\mathcal{N}_x (\mu_r, \Sigma_r)} dx,
\end{align*}
\noindent where $\mathcal{N}_x$ denotes the probability density function of the normal distribution.

\item Semantic matching models exploit similarity-based scoring functions. They measure plausibility of facts by matching latent semantics of entities and relations embodied in their vector space representations. One of the representative algorithms for learning by semantic matching is RESCAL \cite{nickel2011three}. RESCAL optimizes the following expression:
\begin{equation*}
f_r(h,t) = \textbf{r}^{T} \cdot M_r \cdot \textbf{t},
\end{equation*}
\noindent where $\textbf{h}$ and $\textbf{t}$ are representations of entities, and $M_r \in \mathbb{R}^{d \times d}$ is a matrix associated with relations.
\item Matching using neural networks. Deep neural networks  model triplets via training of neural network architectures. One of the first approaches was Semantic Matching Energy (SME) \citep{bordes2014semantic}. This method first projects entities and their relations to their corresponding vector embeddings. The relation's representation is next combined with the relation's head and tail entities to obtain $g_1(\textbf{h},\textbf{r})$ and $g_2(\textbf{t},\textbf{r})$
entity-relation representations in the hidden layer. Finally, a dot product is used to score the triplet relation matching
\begin{equation*}
f_r(h,t) = g_1(\textbf{h},\textbf{r})^{T} \cdot g_2(\textbf{t},\textbf{r}).
\end{equation*}

The simplest version of SME defines the $g_1$ and $g_2$ as:
\begin{align*}
g_1(\textbf{h},\textbf{r}) &= W_1^{(1)} \cdot \textbf{h} + W_1^{(2)} \cdot \textbf{r} + b_{1} \\
g_2(\textbf{t},\textbf{r}) &= W_2^{(1)} \cdot \textbf{t} + W_2^{(2)} \cdot \textbf{r} + b_{2}.
\end{align*}
Here, $W_1^{(1)},W_1^{(2)},W_{2}^{(1)}$ and $W_2^{(2)}$ are $\mathbb{R}^{d \times d}$ dimensional weight matrices and $b_1$ and $b_2$ are bias vectors.
\end{itemize}

Recent advances in embeddings of knowledge graphs show interesting research directions. For example, hyperbolic geometry could be used to better capture latent hierarchies, commonly present in real-world graphs \cite{nickel2017poincare}. Further, KG embedding methods are increasingly tested on large, multi-topic data collections, for example, the Linked Data (LD) which  standardize and fuse data from different resources. Knowledge graph embeddings, such as RDF2vec \cite{rdf2vec} attempt to exploit vast amounts of information in LD and transform it into a learning-suitable format. As knowledge graphs are not necessarily the only source of available information, algorithms  exploit also other information, e.g., textual information available for each triplet \cite{wang2014knowledge}. Recent trends in knowledge graph embeddings also explore how symbolic, logical structures could be used during embedding construction.
Approaches such as KALE \cite{guo2016jointly}  construct embeddings by taking into account logical rules (e.g., Horn clauses) related to the knowledge graph, thus increasing the quality of embeddings. Similar work was proposed by \citet{rocktaschel2015injecting}, where pairs of embeddings were considered during optimization. The same group also showed how relations can be modeled without grounding the head and tail entities for simple implication-like clauses \citep{demeester2016lifted}.
\citet{wang2015knowledge} demonstrated that logical rules can aid in knowledge graph completion on large knowledge bases. They showed that inclusion of rules can reduce the solution space and significantly improve the inference accuracy of embedding models.

\subsubsection{Entity embedding with the StarSpace approach}
\label{sec:entity}


The guiding principle behind all embeddings, described in the previous section, is the persistence of similarity, i.e. that entities which are similar in the knowledge graph must be represented by vectors that are similar in the embedding space. A general approach implementing this principle is to use any similarity function between entities to form a prediction task for a neural network. Below we describe a successful example of this approach, called StarSpace \cite{Wu2018}. As this approach assumes discrete features from a fixed dictionary, it is particularly appealing to relational learning and inductive logic programming.

The idea of StarSpace is to form a prediction task where a neural network is trained to predict the similarity between an entity and its related entity (e.g., its label or some other entity).  
The resulting neural network can be used for several purposes: directly in classification, to rank instances by their similarity, or weights of the trained network can be used as  pretrained embeddings.


In StarSpace, each entity has to be described by a set of discrete features from a fixed-length dictionary and forms a so called Bag-Of-Features. This representation is general enough to cover texts (documents or sentences can be described by bags-of-words or bags-of-n-grams), users (described by  bags of documents, movies, or items they like), relations and links in graphs (described by semantic triples), etc. 
During training, entities of different kinds are embedded \emph{in the same} latent space, suitable for various down-stream learning tasks, e.g., a user can be compared with the recommended items. Note that entities can be embedded along with target classes, resulting in \emph{supervised embedding learning}. This type of representation learning is the key element of the proposed PropStar algorithm outlined in  Section~\ref{sec:featureEmbedding} and presented in detail in Section~\ref{sec:algo-propembed}.

The StarSpace approach trains a neural network model to predict which pairs of entities are similar and which are dissimilar. Two kinds of training instances are formed, positive $(a,b) \in E^+$, which are task dependent and contain correct relations between entities (e.g., document $a$ with its correct label $b$), and negative instances $(a, b_1^-), \dots, (a, b_k^{-}) \in E^-_a$.  For each entity $a$  (e.g., a document) appearing in the positive instances, negative instances are formed using $k$-negative sampling from labels $\{b_i^-\}_{i=1}^k$   as in word2vec \cite{word2vec}. In each batch, the neural network tries to minimize the loss function $L$, defined as follows: 
$$
\textrm{L} = \sum_{(a, b) \in E^+} \bigg( \textrm{Loss} ( \textrm{sim}(a,b)) + \frac{1}{k}\sum_{i=1 \atop (a, b^-_i) \in E^-_a}^k \textrm{Loss} ( \textrm{sim}(a,b^-_i)) \bigg )
$$

%
For each batch update in the training of neural network, $k$ negative examples (a parameter) are formed by randomly sampling labels $b^-_i$ from within the set of entities that can appear in $b$. For example, in the document classification task, document $a$ has its correct label $b$, while $k$ negative instances have their labels $b^-_i$ sampled from the set of all possible labels. Similarity function \emph{sim} represents the similarity between the vector representations of the two entities; typically a dot product similarity is used.
Within one batch, loss function $\textrm{Loss}$ sums the losses of the positive instance $(a,b)$ and the average of the $k$ negative instances $(a,b^-_i), i \in 1\dots k$. To asses the loss, margin ranking loss is used,
$\textrm{Loss} = \max(0, m - \textrm{sim}(a, b'))$, where $m$ is the margin parameter, i.e. the similarity threshold, and $b'$ is a label.

The trained network can be used for several purposes. To classify a new instance $a$, one iterates over all possible labels $b'$ and chooses $\argmax_{b'} \mathrm{sim}(a,b')$ as the prediction. For ranking, entities can be sorted by their predicted similarity score. The embedding vectors can also be extracted and used for
some other downstream task. \citet{Wu2018} recommend that the similarity function
$sim(\cdot, \cdot)$ is shaped in such a way that it will directly fit
the intended application, so that training will be more effective.

A few examples of tasks successfully tackled with the StarSpace feature transformation approach are described below.
\begin{itemize}
\item In \emph{multiclass text classification} the positive instances $(a,b)$ are taken from the training set of documents $E^+$, represented with  bags-of-words and their labels $b$.  For negative instances, entities $b_i^{-}$ are sampled from the set of possible labels. 
\item In \emph{recommender systems} users are described with a bag of items they liked (or bought). The positive instances use a single user ID as $a$ and one of the items that user liked as $b$. Negative instances take $b_i^{-}$ from the set of possible items. Alternatively, to work for new users, the $a$ part of user representation is composed of all the items that user liked, except one, which is used as $b$.
\item For \emph{link prediction} the concepts in a graph are represented as triples head-relation-tail $(h, r, t)$, e.g., gene-generates-protein. A positive instance 
$a$ consists either of $h$ and $r$, while $b$ consists of  $t$; alternatively, $a$ consists of $h$, and $b$ consists of $r$ and $t$. Negative instances $b_i^{-}$ are sampled from the set of possible concepts. The trained network can then predicted links, e.g., gene-generates-what.
\item For \emph{sentence embedding} in an unsupervised fashion, a collection of documents,
containing sentences, is turned into a training set. For positive instances, $a$ and $b$ are sentences from the same document (or are close together in a document), while for negative instances, sentences $b^-_i$ are coming from different documents. This definition of a task tries to capture the semantic similarity between sentences in a document. 
\end{itemize}

In the PropStar algorithm proposed in this work, we use StarSpace similarly to the first case mentioned above (multiclass text classification). Namely, Wordification returns a bag of features (relational items) for each instance in the target table. The embeddings are learned for each feature separately, and class labels are also embedded in the same space. During classification, representations of relational items associated with a given instance (bag of features) are averaged to obtain the representation of the instance---a similar idea as in the document representation adopted in the highly efficient doc2vec branch of algorithms aimed at document classification \citep{le2014distributed}. The embedded instances, now located in the same vector space as the embeddings of class labels, are directly used for classification.  The label, closest to the representation of a given target instance is selected as the final prediction.

\subsubsection{Deep Relational Machines}
\label{sec:drm}

Deep neural networks are effective learners in numeric space,  capable of constructing intermediate knowledge constructs and thereby improve semantics of baseline input representation. Training deep neural networks on propositionalized relational data were explored by \citet{srinivasan209jmlr}, following the work of 
\citet{lodhi2013deep}, where Deep Relational Machines (DRMs) were first introduced. In Lodhi's work, the DRMs used bodies of first order Horn clauses as input to restricted Boltzmann machines, where conjuncts of bonds and other molecular structure information compose individual complex features; when all structural properties are present in a given instance, the target's value is true, and false otherwise. For example, consider the following propositional representation of five instances (rows),  where complex features are comprised of conjuncts of atoms $f_i$, as illustrated in Figure~\ref{fig:example-drm}.

\begin{figure}[h]
\centering

\begin{tabular}{ccccccc}
\rotatebox{70}{Instance} &
\rotatebox{70}{$f_1 \wedge f_2$} & \rotatebox{70}{$f_3 \wedge f_2$} & \rotatebox{70}{$f_1 \wedge f_3$} & \rotatebox{70}{$f_5 \wedge f_2$} & \rotatebox{70}{$f_4 \wedge f_1 \wedge f_5$} & \rotatebox{70}{~Class} \\
  1  & [~1 & 1 & 1 & 1 & 0~] & \textbf{\textcolor{-red!75}{+}} \\
  2  & [~0 & 1 & 0 & 0 & 1~] & \textbf{\textcolor{-red!75}{+}}\\
  3  & [~0 & 0 & 1 & 0 & 0~] & \textbf{\textcolor{red!75}{-}} \\
  4  & [~0 & 1 & 0 & 0 & 1~] & \textbf{\textcolor{red!75}{-}} \\
  5  & [~1 & 0 & 0 & 0 & 1~] & \textbf{\textcolor{red!75}{-}} \\
\end{tabular}
\caption{Example input data for a deep relational machine that operates on the instance level.}
\label{fig:example-drm}
\end{figure}


\indent Note that the propositionalized data set $P$ is usually a sparse matrix, which can represent additional challenge for neural networks.
The DRMs proposed by \citet{lodhi2013deep} were used for prediction of protein folding properties, as well as mutagenicity assessment of small molecules. This approach used feature selection with information theoretic measures such as information gain as the sparse matrix resulting from the propositionalization was not suitable as an input to the neural network. 
The initial studies regarding DRMs explored how deep neural networks could be used as an extension of relational learning. 

Recently, promising results were demonstrated in the domain of molecule classification \cite{dash2018large} using ILP learner Aleph in its propositionalization mode for feature construction. After obtaining propositional representation of data, the obtained data table was fed into a neural network that associated such representations with the output space (e.g., a molecule's activity). Again, sparsity and size of the propositionalized representation is a problem for deep neural networks. Again,  stochastic feature selection of relational features that are used as input to deep relational machines can improve the performance and interpretability \cite{dash2019}. 

The work of
\citet{srinivasan209jmlr} is relevant for the interpretability of deep relational machines, proposing a logical approximation of well-known prediction explanation method LIME \cite{Ribeiro16} and showing how it can be efficiently computed. 

In summary, DRMs address the following issues at the intersection of deep learning and relational learning:
\begin{itemize}
\item DRMs demonstrated that deep learning on propositionalized relational structures is a sensible approach to relational learning.
\item Their input is comprised of logical conjuncts, offering the opportunity to obtain human-understandable explanations.
\item DRMs were successfully employed for classification and  regression.
\item Emerging ideas in the area of representation learning have only recently been explored in the ILP context \cite{dumancic2018auto}, indicating there are many possible improvements both in terms of execution speed, as well as more informative feature construction on the symbolic side of computation.
\end{itemize}

We further discuss DRMs in the context of efficiency of their implementation in Sections~\ref{sec:instanceEmbedding} and \ref{sec:algo-drm}.
 Development of DRMs that are efficient with respect to both space and time is an ongoing research effort.  Building on the ideas of DRMs, we implemented a variant of this approach, capable of learning directly from large, sparse matrices that are returned from Wordification of a given relational database, rather than using feature selection or the output of Aleph's feature construction approach. Our novel, efficient DRM implementation is presented in Section~\ref{sec:algo-drm}.


\section{Unifying framework for propositionalization and embeddings}
\label{sec-unifyingFramework}

The connection we made between different information representation levels and different transformation techniques shows that propositionalization and embeddings are two sides of the same coin. If we view embeddings as transformations for  texts, graphs, recommendations, electronic health records, and other entities with defined similarity function, we can conclude that all these transformation present a multifaceted approach to feature construction. 

To this end, the paper contributes a novel understanding of these data transformation techniques. In Section \ref{sec:unifiedDefinitions}, we first present a unified terminology and definitions, and explain the apparent differences between the definitions of propositionalizationa and embeddings as variants of a  complex data transformation task. In further sections we 
explore the apparent differences between the two approaches.
In Sections \ref{sec:unifiedRepresentation}, \ref{sec:unifiedLearning}, and \ref{sec:unifiedUse}  we discuss  differences in data representation, learning, and use. Finally, in Section \ref{sec-summarylimitations} we summarize strengths and limitations of propositionalization and embeddings.

\subsection{Unifying definitions}
\label{sec:unifiedDefinitions}
Below we present a unified view on the definitions of propositionalization and embedding tasks, as instances of a general data transformation task defined in Section~\ref{def-transformation} via
Definition~\ref{def-transformation}.

\begin{definition}[Propositionalization]
\label{def-propositionalization}

\begin{description}
\item[{\bf Given:}] Input data of a given data type and format, and heterogeneous background knowledge of various data types and formats.
\item[{\bf Find:}] A tabular representation of the data enriched with the background knowledge, where each row represents a single data instance, and each column represents a feature in a d-dimensional symbolic\footnote{In the case of binary valued features, each value in each column is $\in \{0,1\}$.} vector space \cal{F}$^d$.
\end{description}
\end{definition}

\begin{definition}[Embedding]
\label{def-embedding}

\begin{description}
\item[{\bf Given:}] Input data of a given data type and format, and heterogeneous background knowledge of various data types and formats.
\item[{\bf Find:}] A tabular representation of the data enriched with the background knowledge, where each row represents a single data instance, and each column represents one of the dimensions in the d-dimensional numeric vector space $\mathbb{R}^d$.
\end{description}
\end{definition}

%

\subsection{Unifying propositionalization and embeddings in terms of data representation}
\label{sec:unifiedRepresentation}

Both data transformation techniques result in a vector space representation. The unifying dimensions of propositionalization and embeddings in terms of data representation, which are summarized in Table~\ref{tab-dataRepresentation}, are explained below.

In propositionalization, the transformation results in a binary matrix of sparse binary vectors, where rows corresponds to training instances and columns correspond to symbolic features constructed by a particular propositionalization algorithm. These features are human interpretable, as they are either simple logical features (such as attribute values), conjunctions of such features, relations among simple features (such as e.g., a test for the equality or inequality of values of two attributes of the same type), or relations among entities (such as links among nodes in a graph). Given that the number of constructed features is usually large, such transformation results in a sparse binary matrix with few non-zero elements.

Embeddings output is usually a dense matrix of a user-defined dimensionality, composed of vectors of numeric values, one for each object of interest. For neural network based embeddings, vectors usually represent the activation of neural network nodes of one or more levels of a deep neural network. Given a relatively low dimensionality of these vectors (from 100 to 1000) this dense representation is efficient in terms of space. However, the features/dimensions are non-interpretable, therefore a separate explanation mechanisms and visualizations are required.

\begin{table}[htb]
\centering
\caption{Unifying and differentiating aspects of propositionalization and embeddings in terms of data representation.
\label{tab-dataRepresentation}}
\begin{tabular}{lll}
\hline
Representation  & Propositionalization & Embeddings\\\hline
Vector space & symbolic & numeric  \\
Features/variables      & symbolic   & numeric  \\
Feature values          & Boolean (0 or 1) & numeric \\
Sparsity               & sparse            & dense \\
Space complexity & space consuming & mostly efficient \\
Interpretability & interpretable features & non-interpretable\\
 \hline
\end{tabular}
\end{table}

\subsection{Unifying propositionalization and embeddings in terms of learning}
\label{sec:unifiedLearning}

For both data transformation techniques, the resulting vector space representation is used as an input to a learning algorithm of the user's choice. The unifying dimensions of propositionalization and embeddings in terms of most frequently used learners (summarized in Table~\ref{tab-learning}) are explained below.

After propositionalization, any learner capable of processing symbolic features can be used. Typical learners include rule learning, decision tree learning, random forests for a supervised setting, or association rules and symbolic clustering algorithms applied in a non-supervised learning setting. Learners usually use heuristic search to find a global optimum in terms of heuristics to be optimized (exceptions being, e.g., association rule learners using exhaustive search with constraints). Typical algorithms are decision tree learners, rule learners, linear regression and SVMs. Learners require some parameter tuning to achieve optimal results, but parameters are relatively few. Learning is typically performed on CPUs.

The embedded vectors are best suited for distance-based learners, such as neural networks, and to a lesser degree for kernel methods or logistic regression. Deep neural networks use greedy search to find  locally optimal solutions, and are usually trained on GPUs, but can be used for prediction on both CPUs or GPUs. As a weakness, deep learning algorithms require substantial (hyper)parameter tuning.

\begin{table}[ht]
\centering
\caption{Unifying and differentiating aspects of propositionalization and embeddings in terms of learning context.\label{tab-learning}}
\begin{tabular}{lll}
\hline
Learning & Propositionalization & Embeddings\\\hline
Meaning capturing & via symbols &  via distances \\
Search strategy &  heuristic search & greedy \\
Search goal     & global optimum & local optimum \\
Typical algorithms & symbolic, linear regression, SVM   & deep neural networks \\
Parameters      & few  & many  \\
Hardware & CPU & CPU/GPU \\
 \hline
\end{tabular}
\end{table}

\subsection{Unifying propositionalization and embeddings in terms of use}
\label{sec:unifiedUse}

The unifying dimensions of propositionalization and embeddings in terms of their use (summarized in Table~\ref{tab-use}) are explained below.

\begin{table}[ht]
\centering
\caption{Unifying and differentiating aspects of propositionalization and embeddings in terms of use. \label{tab-use}}
\begin{tabular}{llll}
\hline
Use  & Propositionalization & Embeddings\\\hline
Problems/context  & relational & tabular, texts, graphs \\
Data type fusion & enabled & enabled   \\
Explanation & directly interpretable & special approaches \\
 \hline
\end{tabular}
\end{table}

Propositionalization \cite{Propositionalization} is one of the established methodologies used in  relational learning \cite{rdm2001,deraedtbook:2008} and ILP \cite{ILP-book,ILP-Lavrac,deraedtbook:2008} (see the  propositionalization methods outlined in Section~\ref{sec:propositionalization}). The propositionalization approach was applied also in the semantic data mining where ontologies are used as a background knowledge in relational learning \cite{segmine_bmc,SemanticSD.2009,gsegs}. 

The embedding technologies are mostly used in the context of deep learning for various data formats, including tabular data, texts, images, and graphs (including knowledge graphs).
In addition to knowledge graph embedding approaches (see Section~\ref{sec:kge}), we outline some other approaches to graph embeddings below.

The first studies of graph embeddings were influenced by embedding construction from textual data. For example, the well known skip-gram model, initially used as part of word2vec \citep{word2vec} was successfully applied to learn node representations. DeepWalk \citep{deepwalk} was one of the first learners that treats short random walks in graphs as sentences (or short phrases) to learn latent node embeddings. DeepWalk was revisited as node2vec \cite{node2vec} to take into account different types of random walks, parameterized by breadth, as well as depth-first search. LINE \cite{LINE} performs similarly well for the tasks of classification and link prediction by attempting to optimize both local, as well as global network structure. 

As for fusing heterogeneous data types, a propositionalization approach was proposed as a mechanism for heterogeneous data fusion \cite{grcar2012}. As for data type fusion using embedding-based methods, PTE \cite{tang2015pte} exploits heterogeneous networks of texts for supervised embedding construction. NetMF \cite{Qiu2018} is a generalization of Deepwalk, node2vec, LINE and PTE, re-formulating them as a matrix factorization problem. Furthermore, struc2vec \cite{struc2vec} builds on two main ideas: representations of two nodes must be close if the two nodes are structurally similar, and the latent node representation should not depend on any node or edge attribute, including the node labels. Examples of approaches to heterogeneous graph embeddings include HINMINE \cite{Kralj2018}, metapath2vec \cite{zhu2018prediction} and OhmNet \cite{zitnik2017predicting}, an extension of node2vec to a heterogeneous biological setting. Heterogeneous data embeddings \cite{chang2015heterogeneous} of images, videos and text were also formulated as a task of heterogeneous graph embedding.

Concerning the interpretability of results, propositionalization approaches are mostly used with symbolic learners whose results can be interpretable, given the interpretability of features used in the transformed data description. For embedding-based methods, given the non-interpretable numeric features/dimensions, specific mechanisms need to be implemented to ensure results explanation \cite{Robnik08TKDE,strumbelj2014explaining}. A recent well-known approach, which can be used in a post-processing phase of an arbitrary prediction model, is named SHAP \cite{shap}. In this approach, Shapley values offer insights into instance-level predictions by assigning fair credit to individual features for participation in prediction-explaining interactions. Explanation methods such as SHAP are commonly used to understand and debug black-box models. We refer the reader to \cite{shap} for a detailed overview of the method. 

\subsection{Summary of strengths and limitations of propositionalization and embeddings}
\label{sec-summarylimitations}

Let us summarize the unified presentation of propositionalization and embeddings by presenting the strengths and weaknesses of the two approaches.
The main strength of propositionalization is the interpretability of the constructed features, while the main strength of embeddings is high performance of classifiers learned from embeddings due to their compact representation in a vector space.

In terms of their strengths, both approaches to data transformation are: (a) automated, (b) fast, (c) semantic similarity of instances is preserved in the transformed instance space (as a remark, due to a more compact representation, embeddings preserve semantic similarity of features even better than propositionalization), (d) transformed data can be used as input to standard propositional learners, as well as to contemporary approaches.

In addition to these characteristics, embeddings have other favorable properties: 
(a) embedded vectors representations allow for transfer learning, e.g., for cross-lingual applications in text mining or image classification from different types of images, (b) cover a very wide range of data types (text, relations, graphs, images, time series), and (c) have a very wide community of developers and users, including industry.


In terms of their limitations when used in a multi-relational setting, both approaches to data transformation: (a) are limited to 1-many relationships (cannot handle many-to-many relationships between the connected data tables), (b) cannot handle recursion, and (c) cannot be used for predicate invention.

In addition to these characteristics, limitations of propositionalization include: (a) only boolean values are used in the transformed vector space, (b) generated sparse vectors can be memory inefficient, (c) limited range of data types are handled (relations, graphs), and (d) a small community of developers and users (mainly from ILP).

Embeddings also have several limitations: (a) loss of explainability of features and consequently of the models trained on the embedded representations, (b) many user-defined hyper-parameters, (c) high memory consumption due to many weights in neural networks, and  (d) requirement for specialized hardware (GPU) for efficient training of embeddings, which may be out of reach for many researchers.


\section{Proposed unification methodology and its two implementations}
\label{sec:implementations}

The unifying aspects analyzed in Section~\ref{sec-unifyingFramework} can be used as a basis for a unifying methodology that combines propositionalization and embeddings, and benefits from the advantages of both.  The propositionalization successfully captures relational information through complex relational feature construction, but results in a sparse symbolic feature vector representation. This weakness can be successfully overcome by embedding the constructed feature vectors into a lower dimensional numeric vector space, resulting in a condensed numeric feature vector representation appropriate for use by modern deep learning algorithms. 

To this end,  we describe two novel data transformation algorithms, combining propositionalization and embedding based transformations into a joint data transformation framework. We first briefly outline the two approaches in Section~\ref{sec:twoPipelinesOverview}, followed by their detailed descriptions in Section~\ref{sec-detailedTransformations}.

\subsection{Outline of proposed data transformation and learning methods}
\label{sec:twoPipelinesOverview}

We first overview the proposed unifying data transformation approaches. The first, named PropDRM, is an instance-based data transformation approach.
The second one is a feature-based data transformation pipeline, called PropStar. 
The approaches are outlined in the next two subsections.

\subsubsection{PropDRM: An instance-based approach}
\label{sec:instanceEmbedding}
The first unifying approach for embedding of multi-relational databases is based on Deep Relational Machines \citep{dash2018large} (DRMs), presented in Section~\ref{sec:drm}. Rather than using the output of Aleph's feature construction approach, as was the case in the DRM implementation of \citet{dash2018large}, we implemented a variant of this approach, capable of learning directly from large, sparse matrices that are returned by the Wordification \citep{perovsek2015} approach to propositionalization of relational databases.  In this work, following the paradigm of propositionalization by Wordification, each instance is described by a bag (a multiset that allows for multiple appearances of its elements) of features of the form \emph{TableName}$\_$\emph{AttributeName}$\_$\emph{Value}.   Wordification treats these simple easily interpretable features  as `words' in the transformed \mbox{Bag-Of-Words} representation. In this work, they represent individual `relational items' and we use the notation $(\textrm{table.name}, \textrm{column.name}, \textrm{value})$.

\begin{figure}[ht]
\centering
\includegraphics[width = .95\linewidth]{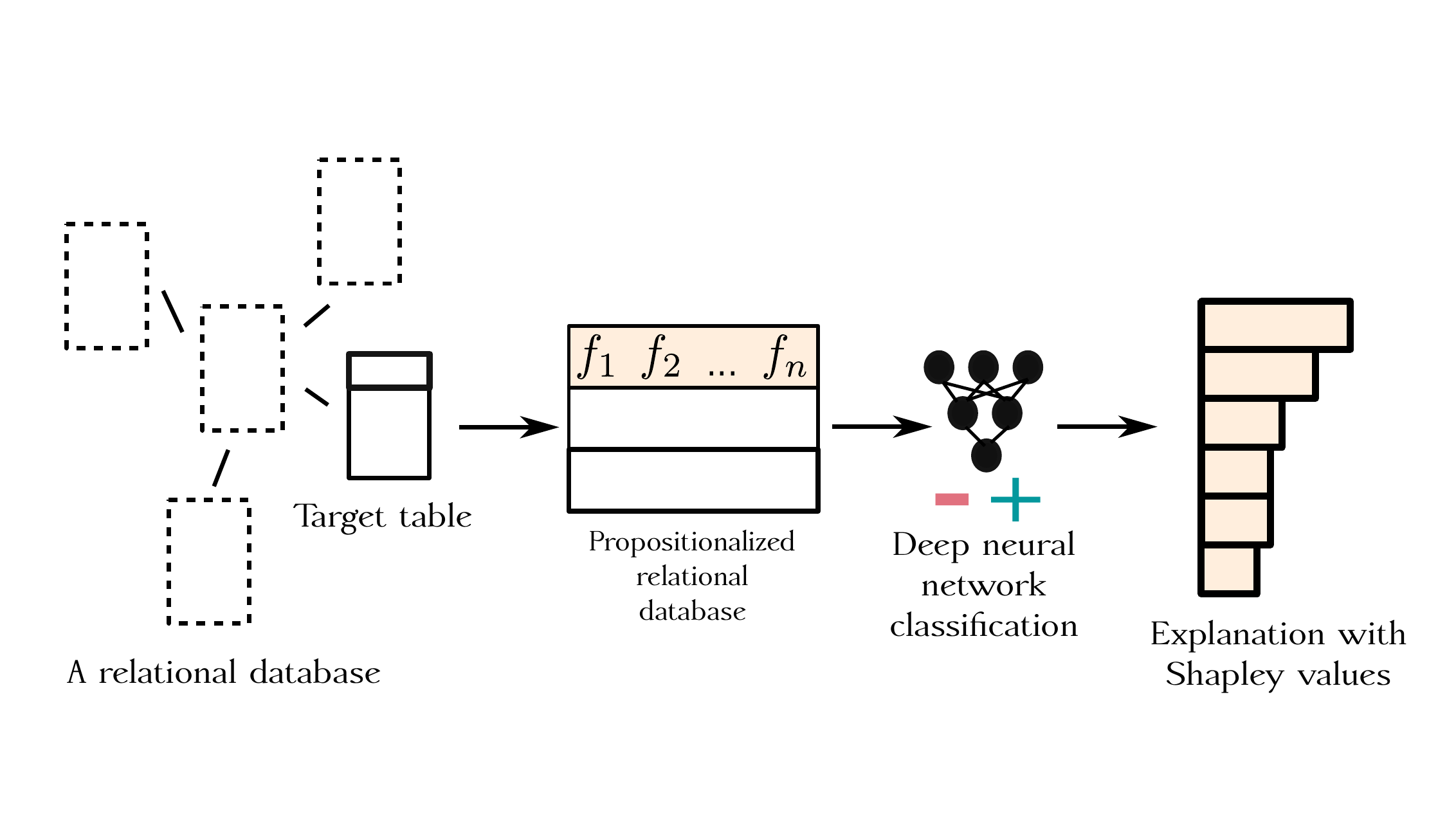}
\caption{Overview of the PropDRM instance-based embedding methodology, based on DRMs. 
Note that features  in the propositionalized relational database represent either single features $f_i$ or conjuncts of features, e.g., $f_i$ $\wedge$ $f_j$, given that Wordifications constructs both feature forms. For simplicity, the propositionalized database shows only two instances.
}
\label{fig:scheme2}
\end{figure}

Relational representations are thus obtained for individual instances, resulting in embeddings of instances (e.g., molecules, persons, companies etc). Batches of instances are then fed to a neural network, which performs the desired down-stream task, such as classification or regression. 
Schematically, the approach is illustrated in Figure~\ref{fig:scheme2}\footnote{As its last step, the methodology includes the explanation of results using the SHAP approach. However, as Section~\ref{sec:implementations} focuses on our research contributions, this well known approach and its results are presented in Appendix~\ref{sec:interpretability}.}.

Note that although propositionalization and subsequent learning are conceptually two distinct steps, they are not necessarily separated when implemented in practice: as neural networks operate with small batches of input data, if propositionalization is capable of similar batch functionality, relational features can be generated in a lazy manner when needed by the neural network. The technical details of the proposed PropDRM implementation are presented in Section~\ref{sec:algo-drm}.

When compared to our PropStar algorithm presented in Sections~\ref{sec:featureEmbedding} and \ref{sec:algo-propembed} below, the key difference of the outlined DRM-based implementation of the unifying methodology is the type of embeddings: PropDRM embeds instances (i.e. whole bags of constructed features), whereas PropStar embeds features along with the class values in the same vector space.

\subsubsection{PropStar: A feature-based approach}
\label{sec:featureEmbedding}

In this section, we outline the proposed PropStar algorithm for classification via feature embedding. Its details and implementation are presented in Section~\ref{sec:algo-propembed}. Unlike the PropDRM algorithm, where each embedding vector represents a single data \emph{instance}, the idea of PropStar is to use embedding vectors to represent the \emph{features} of the data set. 
Here, individual relational features, obtained as the result of propositionalization by Wordification, are used by a supervised embeddings learner to obtain representations, co-located with instance labels. This approach is conceptually different in the sense that representations are not  learned for individual instances (as is the case of DRMs); instead, they are learned for every single relational feature that is the output of the selected propositionalization algorithm (i.e. Wordification). 

The fact that PropStar produces vector representations of features means that the labels (label=true and label=false) are also represented by vectors in the same dense space as the other vectors. This leads to an intuitive direct classification of new examples. We can observe the set of vectors representing the relational items present in the itemset representing the new example. To classify a new instance, the embeddings of the set of its features (i.e. true values) are averaged and the result is compared to the embedding of class labels. The nearest class label is chosen as the predicted value.

\begin{figure}[ht]
\centering
\includegraphics[width = .95\linewidth]{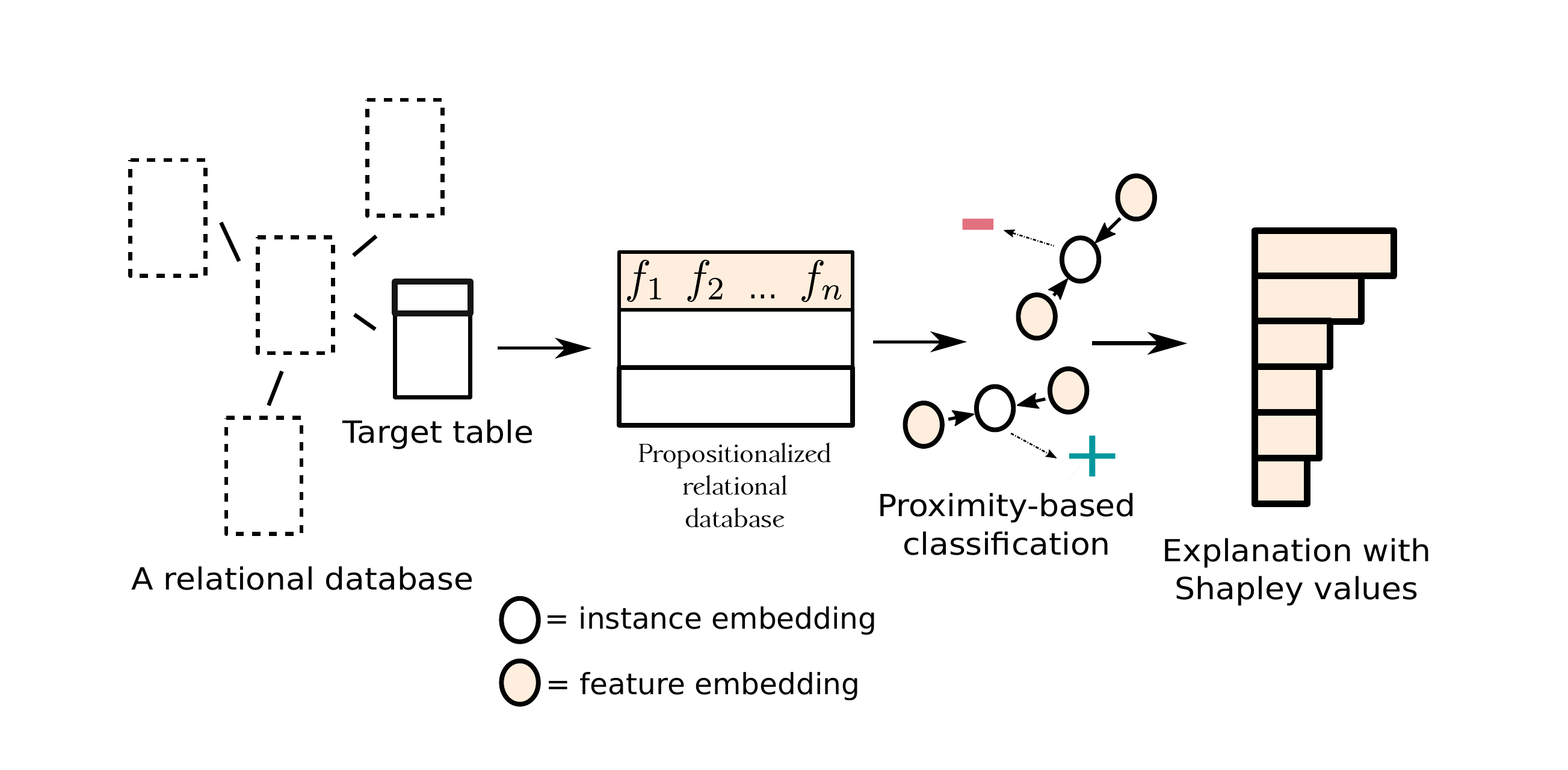}
\caption{Overview of the proposed feature-based embedding methodology PropStar. Note that embedded features represent embeddings of single features $f_i$ or of conjuncts of features, e.g., $f_i$ $\wedge$ $f_j$, given that Wordifications constructs both feature forms. For simplicity, the propositionalized database shows two instances 
Blank and shaded circles correspond to embedded representations of instances and features, respectively. 
}
\label{fig:scheme}
\end{figure}

Figure~\ref{fig:scheme} illustrates how new instances are  classified by direct comparison of the representations of their features in the latent dense semantics-preserving space that also contains the information on labels.  The classification is based on the proximity to a given label (in the latent space). If the center of feature vectors of a given instance is  closer to the vector representing the feature label=true, then the example is classified as positive.

In contrast to the instance-based embeddings discussed in Section \ref{sec:instanceEmbedding}, which relies on batches, the whole data set is needed to obtain representations for individual features. To avoid high spatial complexity, this class of algorithms would ideally operate on sparse inputs. An example of feature-based embeddings are items that are to be recommended to users, where the representation of a given item is obtained by jointly optimizing the item's co-occurrence with other items, as well as other user's properties. In a relational setting considered in this work, we follow the paradigm of propositionalization by Wordification, where each instance is described by a bag of features of the form $(\textrm{table.name}, \textrm{column.name}, \textrm{value})$. Consequently, in the PropStar approach the embeddings represent bags of such features and their conjunctions (of size 2).  There are as many embeddings as there are unique \emph{features} in the propositionalized representation of a given relational database. As such embeddings by themselves do not contain any information which relates them to the desired output space, target values get embedded alongside other features in a supervised manner.

\subsection{Detailed description of proposed data transformation and learning methods}
\label{sec-detailedTransformations}

This section presents the implementations of the proposed methods, preceded by the description of the updates to the Wordification algorithm \cite{perovsek2015} for multi- propositionalization algorithm presented in Section~\ref{sec:wordification-technical}. In Section~\ref{sec:algo-drm} we discuss how Deep Relational Machines (described briefly in Section \ref{sec:drm}), which use neural networks for learning from relational databases, were adapted to operate on sparse matrices generated by an improved Wordification algorithm.
In Section \ref{sec:algo-propembed} we describe a novel algorithm, called PropStar, which  embeds relational features, extracted as part of propositionalization.

\subsubsection{Improving the efficiency of Wordification}
\label{sec:wordification-technical}

 In this work we significantly extend the ideas proposed in Wordification \citep{perovvsek2013wordification,perovsek2015} with the aim to maintain the classification performance, yet improve its \emph{scalability}. Both proposed algorithms build on the idea of Wordification, yet its use in our algorithms is differentiated by the following design decisions:
\begin{enumerate}
\item Inputs do not need to be hosted in relational databases. PropStar operates on .sql files directly. The algorithm supports SQL conventions, as commonly used in the ILP community\footnote{\url{https://relational.fit.cvut.cz/}}. This modification renders the method completely local, enabling offline execution without additional overhead. Such setting also offers easier parallelism across computing clusters.
\item Algorithm is implemented in Python 3 with minimum dependencies for computationally more intense parts, such as the Scikit-learn \cite{pedregosa2011scikit}, Pandas, and Numpy libraries \cite{van2011numpy}.
All database operations are implemented as array queries, filters or similar, unlocking the potential to run PropDRM and PropStar also on GPUs.
\item As shown by \citet{perovsek2015}, Wordification's caveat is extensive sampling of (all) tables. We relax this constraint to close (up to second order) foreign key neighborhood, notably speeding up the relational item sampling part, but with some loss in terms of relational item diversity. For larger databases, minimum relational item frequency can be specified, constraining potentially noisy parts of the feature space.
\end{enumerate}
One of the original Wordification's most apparent problems is its spatial complexity. In this work we address this issue as follows:
\begin{enumerate}
\item Relational items are hashed for minimal spatial overhead during sampling.
\item During construction of the final representation, a sparse matrix is filled based on relational item occurrence.
\item The matrix is serialized directly into list-like structures, suitable for StarSpace algorithm and thus we maintain minimal spatial overhead.
\item Only the final representation is stored as a low-dimensional (e.g., 32) dense matrix.
\end{enumerate}

\subsubsection{Detailed description of the proposed PropDRM implementation}
\label{sec:algo-drm}

The novelty of the proposed implementation of DRM instance-based embedding, inspired by the work of \citet{dash2018large}, concerns its capability to effectively handle the sparseness of the data  with deep neural networks.  The main novelty of the proposed implementation is that it is indeed capable of operating on larger, sparse matrices directly. Such capability is necessary for DRMs to be compatible with propositionalization, which yields large sparse matrices as the main output.
Below we discuss the neural network architecture and its adaptations.

Let $P$ represent a sparse item matrix, as returned by Wordification (discussed in Sections \ref{sec-wordificationOutline} and \ref{sec:wordification-technical}). Note that Wordification is unsupervised, and thus does not include any information on instance labels. The neural network we use (termed $\omega$) represents the mapping $\omega: P \rightarrow \mathcal{C}$, where $\mathcal{C}$ is the set of classes. 
In this work, we experimented with dense feed-forward neural networks, regularized using dropout \cite{srivastava2014dropout}, and ELU activation function \cite{clevert2015fast} (of intermediary weights). The output weights are activated using sigmoid activation ($\sigma$) in order to obtain binary predictions.

\begin{equation*}
ELU(x) = \begin{cases}c(e^{x}-1), & \text{for } x < 0 \\
x & \text{for } x \geq 0\end{cases},
\end{equation*}
\noindent where $c$ is the user-specified constant.
For a given input matrix $P$, an example of a  single hidden-layer neural network is defined as follows.
\begin{align*}
\omega &= \sigma(W_{o}^T (\textrm{ELU}(\textrm{Drop}(W_1^TP + b_1))) + b_{o}).
\end{align*}

\noindent Here, the $\sigma$ is a sigmoid activation, defined as $\sigma(x) = \frac{1}{1 + e^{-x}}$. The $W_1$ is the weight matrix, $P$ the sparse input space, and $b_l$ the bias vector of a given layer $l \in \{0, 1\}$. The described  neural network returned satisfactory results, hence, we did not perform neuroevolution or similar large-scale search for potentially better performing architectures. 
Throughout this work, we use the binary cross-entropy loss, referred to as \emph{Loss}.
For a given probabilistic classifier, which returns a probability $p_{ij}$ of an instance $i$ belonging to a class $j$, the loss function is defined as follows: 
\begin{align*}
\label{eq:loss}
 \textrm{Loss}(i) = \sum_{j\in \mathcal{C}} y_{ij} \cdot \log p_{ij}.
\end{align*}
\noindent Here $y_{ij}$ is a binary value (0 or 1) indicating whether class $j$ is the correct class label assigned to instance $i$, and $\mathcal{C}$ is a set of all the target classes.
In the case of DRMs, where the instances of a relational database (one of the tables) are classified, each of the $|\mathcal{C}|$ output neurons predicts a single probability $p_{ij}$ for a given target class $j \in \mathcal{C}$. If the neural networks are trained in small batches, the results of the $\textrm{Loss}$ function are averaged to obtain the overall loss of a given batch of instances.

Neural networks are adapted for dense inputs such as images and texts, and are not necessarily suitable for large sparse matrices, as considered in this work (i.e. $P$). The proposed variant of DRMs is adapted as follows. Once the batch size $bs$ (a free parameter) is determined, propositionalized representation $P$ is traversed (in chunks of $bs$ instances). Note that each instance is effectively a $d$-dimensional vector. As the neural network operates with dense batches, each  batch is converted to a dense matrix of $bs \cdot d$ elements  that is used during matrix multiplication within the neural network. The spatial complexity is thus at most $\mathcal{O}(bs \cdot d)$. We observed that even by considering batch size of one, the DRMs are stable and efficient.


\subsubsection{Detailed description of the PropStar algorithm}
\label{sec:algo-propembed}
We next present the novel feature-based embedding algorithm that can operate directly on the propositionalized relational databases. The proposed PropStar algorithm merges symbolic and non-symbolic representations as part of a single procedure for obtaining real-valued representations of features in arbitrary relational databases. The pseudocode of the PropStar algorithm is given in Algorithm~\ref{algo:pemb}.

\begin{algorithm}[h]
\KwData{A Relational database $R$, foreign key map $m$}
\Parameter{Entity embedding parameter set $\mathcal{E}$, representation dimension $d$, target table $T$, target attribute $t$}
itemContainer $\leftarrow \textrm{empty bag of items}$  \Comment*[r]{Begin Wordification.}
\ForEach{instance $i$ $\in T$}{
    relationalItems $\leftarrow$ \{\} \;
    candidateKeys $\leftarrow$ getForeignKeys($R$,$i$)  \Comment*[r]{Links to other tables.}
    candidateTables $\leftarrow$ getCandidateTables(candidateKeys, $R$) \Comment*[r]{Linked tables.}
    \ForEach{table $\in$ candidateTables}{
       \While{not final number of items}{
        bagOfItems $\leftarrow$ \textsc{WORDIFY}(table$(m(\textrm{instance}))$) \;
        add bagOfItems to relationalItems \Comment*[r]{Store sampled items.}
       }
    }
    itemContainer[i].add(relationalItems) \Comment*[r]{Store relational items.}
}
relevantFeatures $\leftarrow$ frequencySelection(itemContainer.values, $d$) \; 
symRep$\leftarrow []$ \Comment*[r]{Sparse vector representations of instances.}
\ForEach{instance $i$ $\in$ targetTable}{
        instanceItems $\leftarrow$ itemContainer[i] \; 
        propRep $\leftarrow$ RelationalFeatures(relevantFeatures, instanceItems) \;
        symRep.append(propRep) \;
}
\textit{featureEmbeddings} $\leftarrow$ StarSpace(symRep, $T[t]$, $\mathcal{E}$) \Comment*[r]{Input is a sparse matrix.}
\Return{featureEmbeddings} \;
\caption{The pseudocode of PropStar algorithm.}
 \label{algo:pemb}
\end{algorithm}

The algorithm consists of two main steps. First, a relational database is transformed into sets of features describing individual instances.  The $\textsc{WORDIFY}$ method constructs features of the form $(\textrm{table.name}, \textrm{column.name}, \textrm{value})$ and uses them to describe each individual instance (see Section~\ref{sec:wordification-technical} for a detailed formulation of this step).

Second, sets of relational items (features) are used as input to the StarSpace entity embedding algorithm (described in Section \ref{sec:entity}), producing embeddings for each distinct relational item. 
The StarSpace embeddings are computed using efficient C++ implementation. We wrote a wrapper which seemingly integrates the first part of PropStar (sampling and propositionalization) with the embedding construction. The problem is formulated as a multiclass classification, where the positive pair generator comes directly from a training set of labeled data specifying $(a, b) \in E^+$ pairs where $a$ are relational item `documents' and $b$ are labels (singleton features). Negative entities $b^-_i$ are sampled from the set of possible labels. Inputs can be described as (multi) sets comprised of both relational items $f_i$, their conjuncts, as well as class labels $c_i$. For example,
 \begin{equation*}
 \{f_1, f_2, f_6, f_6 \wedge f_2,c_1\}
 \end{equation*}
\noindent represents a simple input consisting of three relational items, a conjunct and the target label $c_1$. Note that we apply StarSpace in such manner that the representations are learned for \emph{individual relational items}. A representation matrix of dimension $\mathbb{R}^{|W| \times d}$ is produced as the final output ($|W|$ represents the number of unique relational items considered). Intuitively, the embedding construction can be understood as determining relational item locations in a latent space based on their co-occurrence with other items present in all training instances.
The wrapper can be called via `fit' and `predict' methods, commonly used in contemporary data science and machine learning. In this work, we consider the inner product similarity between a pair of vectors $\boldsymbol{e}_1,\boldsymbol{e}_2$ for the construction of embeddings\footnote{Note that $e_1,e_2$ represent vector representations of relational items (i.e. features) in the output of propositionalization.}, i.e.
$\textrm{sim}(\boldsymbol{e}_1,\boldsymbol{e}_2) = \boldsymbol{e}_1^{T} \cdot \boldsymbol{e}_2.$
As the class labels are embedded in \emph{the same} space as individual relational items, classification of novel bags of relational items is possible by direct comparison, as common tasks operating on word embeddings. We discuss this classification below. 

Let $M$ represent a novel instance to be classified. Note that $M$ (without additional index) is considered a multiset of relational items. For prediction purposes, StarSpace averages the representations of relational items present in a given input instance (a bag). The representation is normalized (as during training) and compared to label embeddings in the common space. Representation of a relational bag $\boldsymbol{e}_{M}$ is computed (with considered hyperparameters) as:
\begin{equation*}
\boldsymbol{e}_{M} = \frac{\hadamand\limits_{f_i \in M} \boldsymbol{e}_{f_i}}{\sqrt{|M_{\textrm{unique}}|}},
\end{equation*}
\noindent which is a $d$-dimensional, real-valued vector. 
Note that $\oplus$ in this expression denotes element-wise summation. The $M_{\textrm{unique}}$ represents the set of all (unique) relational features currently considered. Note that original bags of features can be redundant (multisets), yet representations are learned for unique features.
Next, the similarity of this vector is compared to the  label embeddings in the same space. The label that is the most similar to $e_{M}$ is the top-ranked prediction, the second most similar label is the second-ranked prediction, etc. In this work we consider only the top-ranked prediction, resulting in the following label assignment:

\begin{equation*}
\textrm{label}({e_{M}}) = \argmax_{c \in \mathcal{C}} [ \textrm{sim} (\boldsymbol{e}_{M}, \boldsymbol{e}_c) \big ].
\end{equation*}



The complexity of obtaining a single prediction is hence $\mathcal{O}(|\mathcal{C}|)$, not taking the complexity of scalar product for function $\mathrm{sim}$ into account.
The PropStar algorithm first samples the relational items with respect to the target table (lines 2-11 in Algorithm  \ref{algo:pemb}). Binary indicator function (relationalFeatures) is applied to obtain the propositionalized representation of the target table (line 12). Here, zeros represent absence of a given relational items, and ones their presence\footnote{Note that in the actual implementation CSR format of sparse matrices is used to reduce the spatial overhead of storing zeros.}. Finally, StarSpace is used to embed the table into a low-dimensional, real valued embedding (line 19).


The spatial complexity of PropStar is linear with respect to the number of non-zero elements in the propositionalized version of a relational database. The exact spatial complexity can be formulated as follows. Let $\textrm{row}$ represent the average number of rows per table. Let $n_t$ represent the number of tables and $\textrm{col}$ the average number of columns per table. We improve the original spatial complexity of  $\mathcal{O} ( \textrm{rows} \cdot n_t \cdot 2^{\textrm{col}} )$ by introducing a constraint, which determines the maximum number of relational items that can be considered. The exponential term in the initial complexity thus reduces to $\textrm{col}$ times some constant, yielding the complexity of $\mathcal{O} ( \textrm{rows} \cdot \textrm{col} \cdot n_t)$. This formulation yields a scalable propositionalization.

\section{Experimental evaluation}
\label{sec:experimentalEvaluation}

In this section we describe the implementation details of the proposed methods, the relational data sets used in the experiments, and the experimental evaluation of the proposed methods.

\subsection{Implementation and hyperparameters}
We discuss how the proposed methods were implemented, along with the hyperparameters explored. 
Both new methods (PropDRM and PropStar) are implemented in Python, with the following exceptions. In PropDRM, the DRMs are implemented in PyTorch. For PropStar we used the efficient StarSpace implementation written in C++, for which we build a wrapper offering basic embedding training and prediction functionality. 

We used 10-fold stratified cross validation, which was conducted for individual hyperparameter settings. The best setting is reported, other are discussed in ablation studies. Experiments were performed on an of-the-shelf workstation with no GPUs (even though PropDRM and PropStar can exploit them). We intentionally omit the GPU-based training to explore the minimum hardware, required to perform competitively on the selected data sets---ILP baselines, such as Aleph and RSD are Prolog-based, and are to our knowledge not able to use multiple GPU threads simultaneously. The machine on which experiments were conducted had 128GB of RAM and 12 CPUs (Intel i8 series).

In PropDRM,  we varied the dropout rate, learning rate, number of epochs, and the hidden layer size. In PropStar, we varied the number of negative samples, embedding dimension, learning rate, and the number of epochs.

The source code of our implementation is publicly available\footnote{\url{https://github.com/SkBlaz/PropStar}}.

\subsection{Relational data sets}

Five relational database sources\footnote{Freely accessible at \url{https://relational.fit.cvut.cz/}} \cite{motl2015ctu} were used in the experiments. Their characteristics are summarized in Table~\ref{advancedProperties}.

\begin{description}
\item[Trains~\citep{michie1994international}] data set is used in the East-West trains challenge problem, which is well-known in ILP. The East-West trains challenge is to predict whether a train is eastbound or westbound, based on the properties of eastbound and westbound cars.
Trains contain variable number of cars, each having one of various shapes and carrying
various loads.

\item[Carcinogenesis~\citep{srinivasan1997carcinogenesis}] task is to predict carcinogenicity of a diverse set of chemical compounds.  The data set was obtained by testing
different chemicals on rodents, where each trial would take several years
and hundreds of animals.  The data set consists of 329 compounds, of which
182 are carcinogens. 

\item[Mutagenesis~\citep{debnath}] task addresses the problem of predicting mutagenicity of
aromatic and heteroaromatic nitro compounds.  Predicting
mutagenicity is an important task as it is very relevant to the prediction
of carcinogenesis.  The compounds from the data are known to be more
structurally heterogeneous than in any other ILP data set of chemical
structures.  The database contains 230 compounds of which 138 have positive
levels of mutagenicity and are labeled as `active'.  Others have class
value `inactive' and are considered to be negative examples.  We took the
data sets from the original paper~\citep{debnath}, where the data was
split into two subsets: a 188 compound data set and a smaller data set with 42
compounds.  


\item[IMDB] database is publicly available in the SQL format\footnote{http://www.webstepbook.com/supplements/databases/imdb.sql}. This
database contains tables of movies, actors, movie genres, directors, and
director genres.  The data set used in our experiments encompasses only 
movies whose titles and years of production appear in the IMDB's top-250 and
bottom-100 chart (Snapshot taken on July 2, 2012).  The snapshot contains
166 movies, along with all of their actors, genres and
directors.  We designate movies present in the IMDB top-250 chart as
positive examples, and those in the bottom-100 as negatives.
\item[MovieLens] data set from the UC Irvine machine learning repository\footnote{ \url{https://relational.fit.cvut.cz/dataset/MovieLens}}. The data set is similar to IMDB above, however is much larger. Overall, the database consists of more than 1.2 million instances. The task is to predict gender of the movie database's users.
\end{description}
\begin{table}[htb]
\caption{Properties of the experimental data tables.}
\centering
\begin{tabular}{ccc}
\hline 
Trains & \#rows &\#attributes\\\hline
cars & 63 & 10\\
trains & 20 & 2 \\ \hline
Carcinogenesis & \#rows &\#attributes\\ \hline
atom & 9,064 & 5\\
canc & 329 & 2\\
sbond\_1 & 13,562 & 4\\
sbond\_2 & 926 & 4\\
sbond\_3 & 12 & 4\\
sbond\_7 & 4,134 & 4 \\ \hline
Mutagenesis 42 & \#rows &\#attributes\\\hline
atoms & 1,001 & 5\\
bonds & 1,066 & 5\\
drugs & 42 & 7\\
ring\_atom & 1,785 & 3\\
ring\_strucs & 279 & 3\\
rings & 259 & 2 \\ \hline
Mutagenesis 188 & \#rows &\#attributes\\\hline
atoms & 4,893 & 5\\
bonds & 5,243 & 5\\
drugs & 188 & 7\\
ring\_atom & 9,330 & 3\\
ring\_strucs & 1,433 & 3\\
rings & 1,317 & 2 \\ \hline
IMDB & \#rows &\#attributes\\\hline
actors & 7,118 & 4\\
directors & 130 & 3\\
directors\_genres & 1,123 & 4\\
movies & 166 & 4\\
movies\_directors & 180 & 3\\
movies\_genres & 408 & 3\\
roles & 7,738 & 4 \\ \hline
MovieLens & \#rows &\#attributes\\ \hline
actors & 99{,}129 & 3 \\
directors & 2{,}201 & 3 \\
movies & 3{,}832 & 5 \\
movies2actors & 152{,}532 & 3 \\
movies2directors & 4{,}141 & 3 \\
u2base & 946{,}828 & 3 \\
users & 6{,}039 & 4 \\
\hline
\end{tabular}
\label{advancedProperties}
\end{table}

\subsection{Results}
We present the results of the empirical evaluation of the proposed methodologies on the presented set of standard benchmark ILP data sets. The accuracies of individual learners are given in Table~\ref{tbl:res}, and the AUC scores are reported in Table~\ref{tbl:auc}. The results for Aleph, RSD, RelF and Wordification were taken from previous work of \citet{perovsek2015}.

\begin{table}[htb]
\centering
\caption{Classification accuracy on different relational data sets. For the proposed methods, we report average performance over 5 runs. The runs, marked with -- were unable to finish in 12 hours.}
\resizebox{1\textwidth}{!}{
\begin{tabular}{ll|cccccc}
Propositionalization &    Learner &  Carc.  &    IMDB &  Mut188 &  Mut42 &  Trains &  MovieLens\\ \hline
&        MajorityVote & 0.55 &  0.73 & 0.67 & 0.69 & 0.50 & 0.72 \\
       \hline
       Aleph (from \cite{perovsek2015}) &        J48 &          0.55 &     0.73 &          0.60 &         0.69 &    0.55  &  --\\
               Aleph (from \cite{perovsek2015}) &        SVM &          0.55 &     0.73 &          0.60 &         0.69 &    0.70 &  --\\
                 RSD (from \cite{perovsek2015}) &        J48 &          0.60 &     0.75 &          0.68 &         \bf{0.98} &    0.60  &  --\\
                 RSD (from \cite{perovsek2015}) &        SVM &          0.56 &     0.73 &          0.71 &         0.69 &    0.80  &  --\\
                RelF (from \cite{perovsek2015}) &        J48 &          0.60 &     0.70 &          0.75 &         0.76 &    0.65 &  --\\
                RelF (from \cite{perovsek2015}) &        SVM &          0.56 &     0.73 &          0.69 &         0.76 &    0.80 & --\\
       Wordification (from \cite{perovsek2015}) &        J48 &          0.62 &     0.82 &          0.67 &         \bf{0.98} &    0.50 &  --\\
       Wordification (from \cite{perovsek2015}) &        SVM &          0.61 &   0.73 &          0.82 &         0.79 &    0.50 &  --\\ \hline
       Aleph (replicated) & J48 & 0.55 & -- & 0.80 & 0.76 & 0.70 & -- \\
       Aleph (replicated) & SVM & 0.55 & -- & 0.80 & 0.79 & 0.60 & --\\ 
      RSD (replicated) & J48 & 0.56 & 0.84 & 0.88 & 0.92 & 0.60 & --\\
      RSD (replicated) & SVM & 0.60 & 0.82 & 0.89 & 0.84 & 0.80 & --\\ 
       Wordification (replicated) & J48 & 0.47 & \bf{0.85} & 0.91 & 0.88 & \bf{0.90} &  0.60\\
      Wordification (replicated) & SVM & 0.39 & 0.80 & 0.83 & 0.33 & 0.50&  0.72\\
      Treeliker & J48 & 0.58 & -- & 0.77 & 0.81 & 0.50 &  --\\
      Treeliker & SVM & 0.60 & -- & 0.90 & 0.80 & 0.70 &  --\\
      \hline
      PropDRM  & & 0.63 &  0.73  & 0.91  & 0.86 & 0.70 & 0.72 \\
      PropStar &   &          \bf{0.66} &      0.74 &          \bf{0.92} &         0.90 &    0.80 & \bf{0.74}\\
       \hline
\end{tabular}}
\label{tbl:res}
\end{table}

\begin{table}[ht]
\centering
\caption{AUC scores on individual data sets. We report average performance over 5 runs. The runs, marked with -- were unable to finish in 12 hours.
}
\resizebox{1\textwidth}{!}{
\begin{tabular}{ll|cccccc}

Propositionalization &    Learner &  Carc. &     IMDB &  Mut188 &  Mut42 &  Trains & Movies\\ \hline
               Aleph (from \cite{perovsek2015}) &        J48 &          0.50 &     0.50 &    0.68 &         0.50 &    0.55 &  --\\
               Aleph (from \cite{perovsek2015}) &        SVM &          0.50 &     0.50 &    0.68 &         0.50 &    0.70 &  --\\
                 RSD (from \cite{perovsek2015}) &        J48 &          0.59 &      0.59 &          0.54 &         \bf{0.96} &    0.60  &  --\\
                 RSD (from \cite{perovsek2015}) &        SVM &          0.52 &      0.50 &          0.58 &         0.50 &    0.80 & --\\
                RelF (from \cite{perovsek2015}) &        J48 &          0.59 &      0.66 &          0.68 &         0.68 &    0.75 &  --\\
                RelF (from \cite{perovsek2015}) &        SVM &          0.52 &     0.50 &          0.54 &         0.62 &    0.75 &  --\\
       Wordification (from \cite{perovsek2015}) &        J48 &          0.61 &       \bf{0.75} &          0.55 &         0.96 &    0.95 & --\\
       Wordification (from \cite{perovsek2015}) &        SVM &          0.58 &   0.50 &          0.78 &         0.65 &    0.50 &  --\\ \hline
       Alpeh (replicated) & J48 & 0.50 & -- & 0.71 & 0.72 & 0.70 &  --\\
       Aleph (replicated) & SVM & 0.50 & -- & 0.75 & 0.73 &  0.60 &  --\\ 
       RSD (replicated) & J48 & 0.55 & 0.71 & 0.87 & 0.92 & 0.60 &  --\\
       RSD (replicated) & SVM & 0.58 & 0.65 & 0.90 & 0.73 & 0.80 &  --\\
       Wordification (replicated) & J48 & 0.48 & 0.72 & \bf{0.90} & 0.86 & 0.90 &  0.52\\
       Wordification (replicated) & SVM & 0.42 & 0.62 & 0.81 & 0.50 & 0.50  & 0.50\\
       Treeliker & J48 & 0.58 & -- & 0.75 & 0.71 & 0.50 &  --\\
       Treeliker & SVM & 0.58 & -- & 0.88 & 0.68 & 0.70 &  --\\ \hline
      PropDRM  & & \bf{0.63} &  0.68  & \bf{0.90}  & 0.87 & 0.80 &  0.54 \\                 PropStar &   &          \bf{0.63} &     0.66 &          0.87 &         0.87 &    \bf{0.95} & \bf{0.56}\\
 \hline
\end{tabular}}
\label{tbl:auc}
\end{table}


It can be observed that the proposed unifying  approaches perform competitively on most data sets. We can observe a distinct difference in performance on the Mutagenesis data sets, where both PropDRM as well as PropStar do not outperform the baselines on the smaller data set (Mut42), yet notably outperform the (best) baselines on the larger one (Mut188). Further, minor improvement over the baseline algorithms is also achieved on Carcinogenesis data set.

In terms of spatial complexity, the proposed methodology greatly outperforms the alternatives under a given set of constraints. Only PropDRM and PropStar scale to very large relational databases without specialized hardware. Detailed studies regarding the sensitivity of PropDRM and PropStar to their parameters are discussed in Appendices~\ref{appendix-ablation-drm} and~\ref{appendix:ablation-dp}, respectively.

We consider the presented results as very favorable for the two proposed approaches. In particular, PropStar is better than current state-of-the-art methods on 3 out of 6 data sets, and is therefore a method to take into consideration when attempting to solve any new relational problem.




\subsubsection{Study of propositionalization projections}
\begin{figure}[ht]
\centering
\captionsetup{width=\linewidth}
\begin{tabular}{cc}
\subcaptionbox{Mutagenesis 188.}{\includegraphics[width = 2.4in]{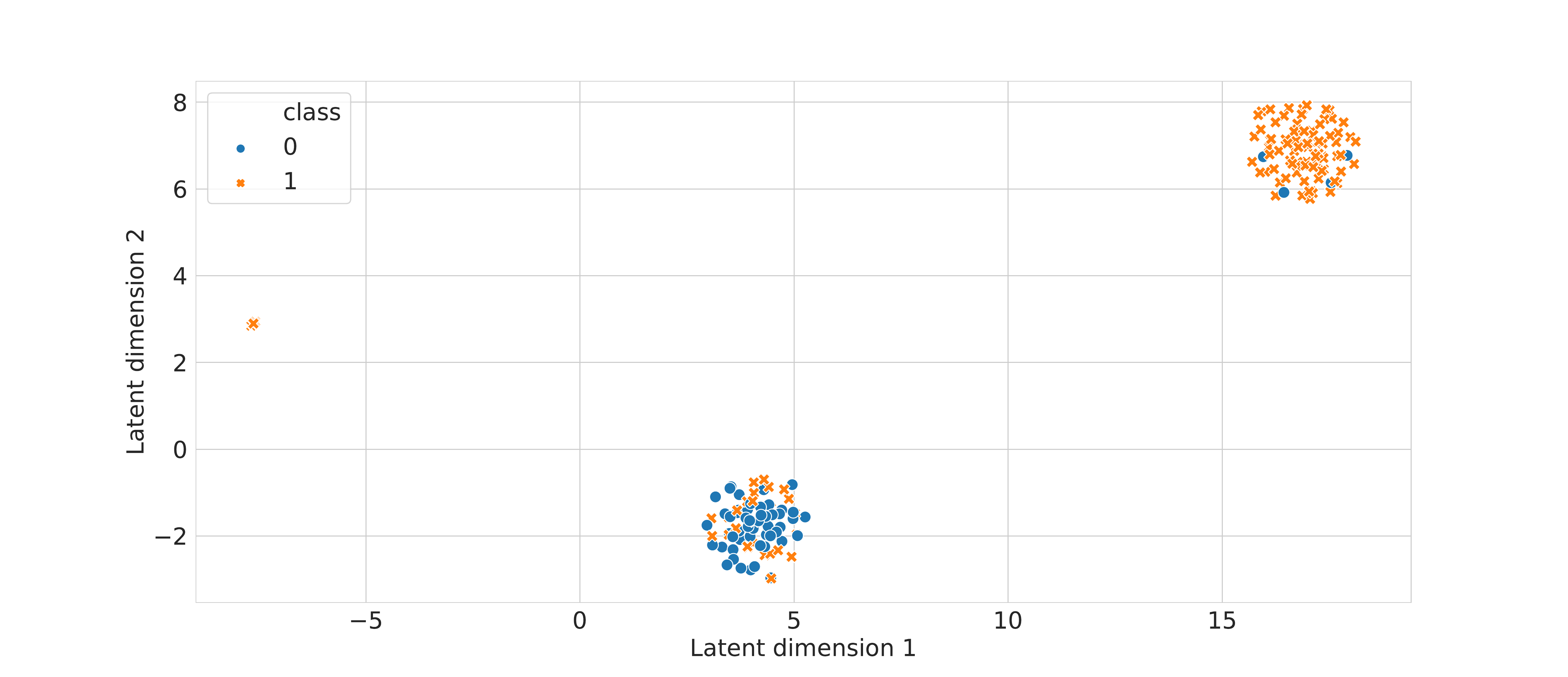}} &
\subcaptionbox{Trains.}{\includegraphics[width = 2.4in]{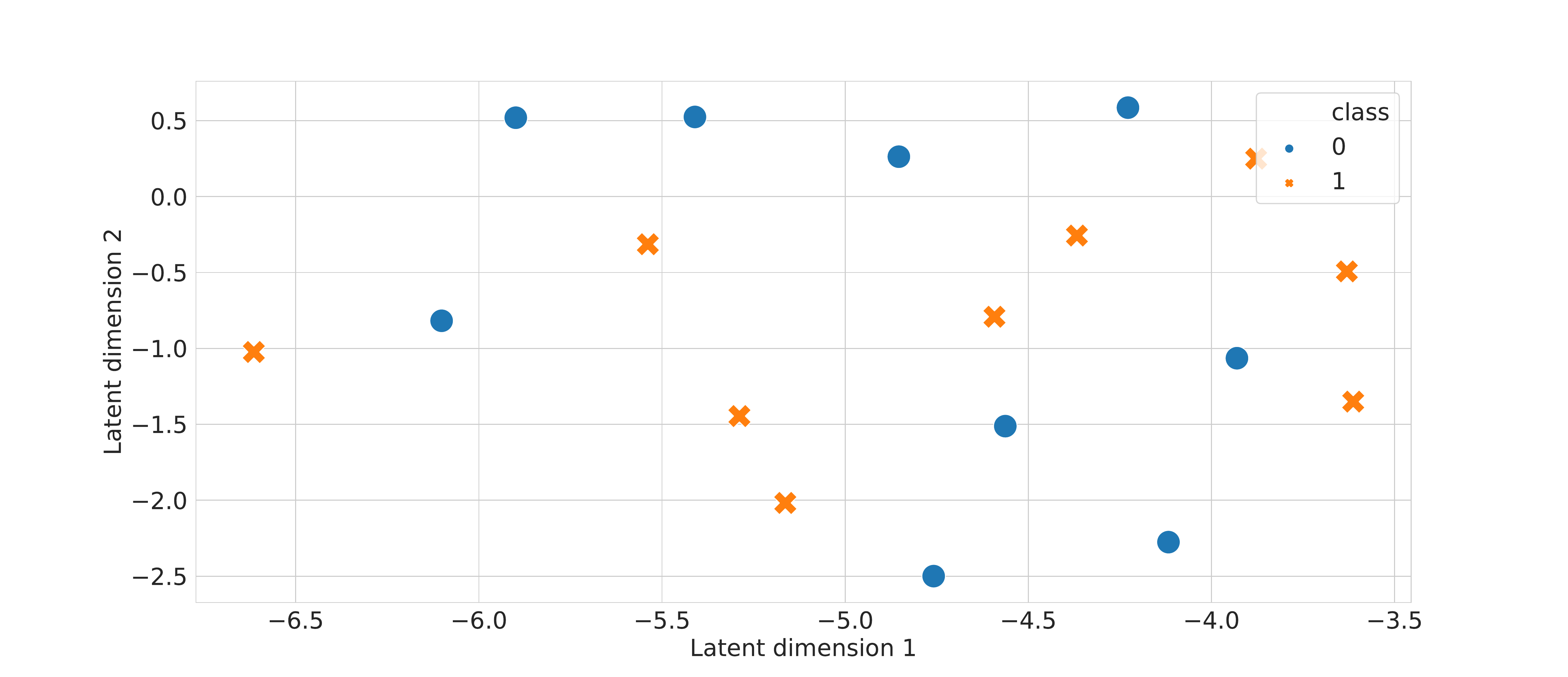}} \\
\end{tabular}
\caption{Two UMAP projections of selected propositionalized data sets.}
\label{fig:umap}
\end{figure}

The considered propositionalization is entirely \emph{unsupervised}. Only once the symbolic representations of instances are obtained,  PropDRM and PropStar learn the associations to individual classes. A good representation, however, already contains relevant information on the instance space. In Figure \ref{fig:umap}, we projected the propositionalized Mutagenesis 188 and Trains instance space to two dimensions to qualitatively explore whether instances group or any meaningful patterns emerge. 
Understanding whether the symbolic space exhibits distinct structure on its own could offer insights into why the proposed methods perform well. For projecting the 10{,}000 dimensional space to two dimensions we used UMAP, a recently introduced non-linear dimensionality reduction method based on insights from manifold theory \cite{mcinnes2018umap-software}. 

We can observe an apparent distinction in the clustering of the UMAP projections of the two propositionalized data sets. The Mutagenesis 188 data set consists of two distinct clusters that, when colored according to the class labels, approximately correspond to the two classes (Figure \ref{fig:umap}a). On the other hand, the clustering is not apparent in the case of the Trains data set (Figure \ref{fig:umap}b), where the instances do not group distinctly. The purpose of the considered visualizations is twofold. First, we show how the symbolic space can exhibit clustering properties, related to properties of instances such as class labels. Next, we show that projections do not necessarily exhibit such properties, indicating potentially harder classification problems. We believe that UMAP and similar tools offer insights into representation structure.

\subsubsection{Statistical comparison of PropDRM and PropStar}
\label{sec:comparison}
In previous sections, we demonstrated that both PropDRM and PropStar perform well on the considered data sets, indicating that both approaches are successfully unifying propositionalization  and embeddings. We further study the differences in performances of the two approaches. For this purpose, we employ the hierarchical Bayesian t-test, a Bayesian test capable of comparing a pair of classifiers across multiple data sets  \cite{bayesiantests2016,corani2017statistical}. For this comparison, we selected the overall best performing hyperparameter sets for each method, and conducted ten repetitions of stratified ten-fold cross validation (for each data set). 
The results are visualized as probability distributions across the space of both classifiers and the `rope' region (region of practical equivalence) within which the two classifiers perform the same. The size of this region is a free parameter of the hierarchical t-test, and was set to 0.05 in this work. 
Other parameters of the test were left as defaults. The exact methodology for the interested reader is explained by \citet{bayesiantests2016}.

In terms of AUC, the probabilities returned by the Bayesian test were as follows:
p(PropStar) = 0.07 and p(PropDRM) = 0.54), and in terms of classification  accuracy, p(PropStar) = 0.96 and p(PropDRM) = 0.04.
The results of statistical analysis indicate that with respect to AUC performance, the two approaches perform similarly, even though the probability that PropDRM will outperform PropStar is higher.
With respect to the classification accuracy, PropStar outperforms PropDRM in majority of comparisons. Thus, considering the 95\% or higher as the probability denoting significance boundary, we can determine that PropStar is (significantly) more suitable choice if accuracy is being optimized for. As Bayesian comparisons are computationally expensive, we compared the two methods using default hyperparameter settings. The PropStar's default configuration is not necessarily optimal when AUC is considered.

\section{Conclusions and further work}
\label{sec-conclusions}
This paper first provides a critical survey of propositionalization and embedding techniques, especially relevant for relational learning and inductive learning programming. While both data approaches,  propositionalization and embeddings, aim at transforming data into the tabular data format, the research papers describing the approaches use different terminology and task definitions, claim to have different goals, and are used in very different contexts. In this paper, we define the main categories of data transformation techniques based on the representation they use and approaches employed. Propositionalization approaches produce tabular data from multirelational databases as well as from a mixture of tabular data and background knowledge in the form of logic programs or networked data, including ontologies. Knowledge stored in graphs can be assessed with community detection and graph traversal methods. Relations described with similarity matrices are encoded in a numeric form using matrix factorization. Currently, the most promising approach to data transformations are neural networks based methods which can be applied to text, graphs, and other entities for which we can define a suitable similarity function.

One of the main strategic problems machine learning has to solve is better integration of knowledge and models across different domains and representations. While the research area of embeddings can unify different representations in a numeric space, symbolic learning  may be an essential ingredient for integration of different knowledge areas. We see our PropStar approach that combines  advantages of propositionalization and neural embeddings in the same data fusion pipeline as a step in that direction.

The first minor contribution of the paper is that our exposition is based on three cognitive representation levels introduced by \citet{Gardenfors:2000}, i.e. neural, spatial, and symbolic. As most of human knowledge is stored in the symbolic form, while the most  powerful machine learning algorithms take as input spatial representations, this explains a plethora of techniques that transform other forms of human knowledge into the spatial representation space. The next contribution is the unifying framework in which we describe propositionalization and embedding techniques in terms of their joint properties and their differences.  We show how the propositionalization techniques can be merged with deep neural network based embedding to produce a joint embedding, such that spatial representation can be used with any deep learning algorithm and the predictions can be comprehensively explained. The main contributions of our work are thus the two implementations that merge propositionalization and embeddings in the same unifying methodology. The first is an efficient reimplementation of existing Deep Relational Machines, while the second one is the novel Deep Propositionalization algorithm. We  also contribute  an experimental evaluation of the two algorithms and show favorable results in terms of predictive performance, as well as time and space requirements.
The source code of both algorithms, DeepProp and PropDRM, is publicly available\footnote{\url{https://github.com/SkBlaz/PropStar}}.

In further work, it is worth investigating the integration of symbolic and deep learning, considering them as multitask learning approaches which try to integrate many different learning tasks and use embeddings as input representations.   
The issue is that different embedding methods have so far only been used in isolation. We already address this challenge in the current work of the authors, where we combine complementary embedding methods from different classes: in particular, to use network traversal methods to produce initial embeddings that are then refined using a deep neural network \citep{DNRSkrlj}.

\begin{acknowledgements}
We acknowledge the financial support of the Slovenian Research Agency through core research programmes P2-0103 and P6-0411 and project \emph{Semantic Data Mining for Linked Open Data} (financed under the ERC Complementary Scheme, N2-0078). The authors have received funding also from the European Union’s Horizon 2020 research and innovation programme under grant agreement No 825153 (EMBEDDIA). The work of the second author was funded by the Slovenian Research Agency through a young researcher grant. We wish to thank Jan Kralj for his insightful comments on the formulation of the proposed framework and for mathematical proofreading.
Further, we are grateful to Vid Podpe\v{c}an and Nika Er\v{z}en for their help with the implementation of the new version of the PyRDM library.
Finally, we would like to thank the anonymous reviewers for careful reading, many insightful observations, and the encouragements to expand the initial work. 
\end{acknowledgements}
\bibliography{bibliography} 

\appendix

\section{Wordification example}
\label{sec-wordificationExample}

The Wordification approach is illustrated on a modified and substantially simplified version of the well-known East-West Trains domain~\citep{michie1994international}. Our input database consists of just two tables shown in Figure~\ref{vlakci}, where we have only one east-bound and one west-bound train, each with just two cars with certain properties\footnote{Note that in the experiments we use the standard version of the East-West Trains domain.}. 

\begin{figure}[hbt]
\begin{center}
\begin{minipage}{0.85\textwidth}
\begin{tabular}{ll}
\multicolumn{2}{l}{\textbf{TRAIN}} \\
\hline
trainID & eastbound \\
\hline
t1 & east \\
\ldots & \ldots \\
t5 & west \\
\ldots & \ldots \\
\hline
\end{tabular} 
\hfill
\begin{tabular}{llll}
\multicolumn{4}{l}{\textbf{CAR}} \\
\hline
carID & shape & roof  & train \\
\hline
c11 & rectangle  & none & t1 \\
c12 & rectangle & peaked   & t1 \\
\ldots & \ldots & \ldots  & \ldots \\
c51 & rectangle  & none  & t5 \\
c52 & hexagon    & flat  & t5 \\
\ldots & \ldots & \ldots  & \ldots \\
\hline
\end{tabular}
\end{minipage}
\caption{Example input for Wordification in the East-West Trains domain.}
\label{vlakci}
\end{center}
\end{figure}

The TRAIN table is the main table and the trains are the instances, with a class label denoting the direction of the train (east of west). A multiset (a bag) of features is generated for each train t1 and t5 with the class label appended  to the resulting feature vector (bag of features). 
Both single features and conjunctive features are shown in this example.

\begin{figure}[htb]
\begin{center}
\small
\begin{verbatim}
t1: {car_shape_rectangle, car_roof_none, car_shape_rectangle__car_roof_none,  
     car_shape_rectangle, car_roof_peaked, car_shape_rectangle__car_roof_peaked}, east
    ...
t5: {car_shape_rectangle, car_roof_none, car_shape_rectangle__car_roof_none,  
     car_shape_hexagon, car_roof_flat, car_shape_hexagon__car_roof_flat}, west
    ...
\end{verbatim}
\caption{\label{docs}The database from Figure~\ref{vlakci} in the Bag-Of-Features representation (as in the original Wordification implementation, conjunctions of features are denoted by a long underscore instead of $\wedge$).}
\label{vlakciProp}
\end{center}
\vspace*{-0.4cm}
\end{figure}

\section{Ablation study---PropDRM}
\label{appendix-ablation-drm}

\begin{figure}[ht]
\centering
\captionsetup{width=\linewidth}
\begin{tabular}{cc}
\subcaptionbox{Dependence on hidden layer size.}{\includegraphics[width = 2.2in]{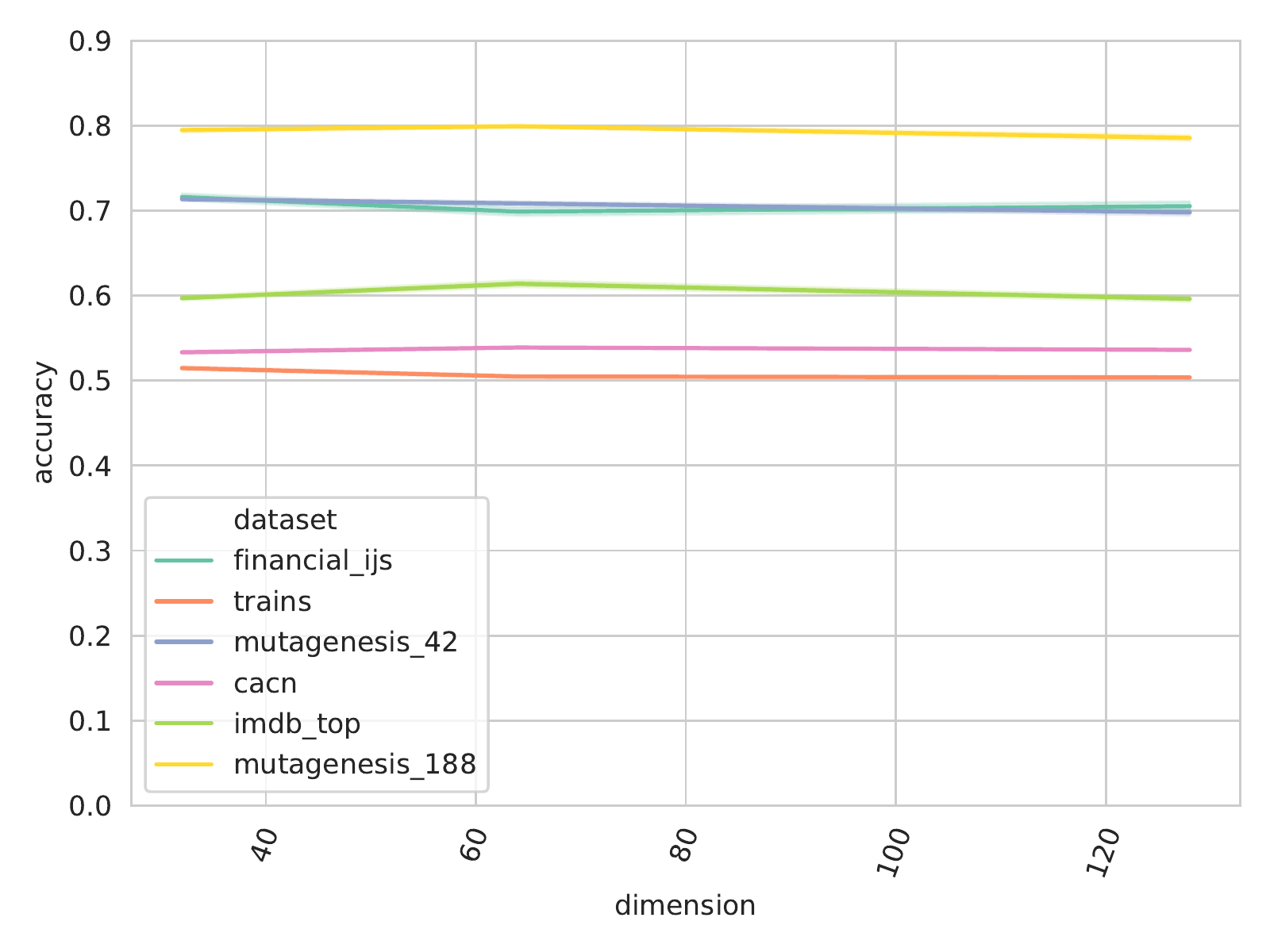}} &
\subcaptionbox{Dependence on the number of epochs.}{\includegraphics[width = 2.2in]{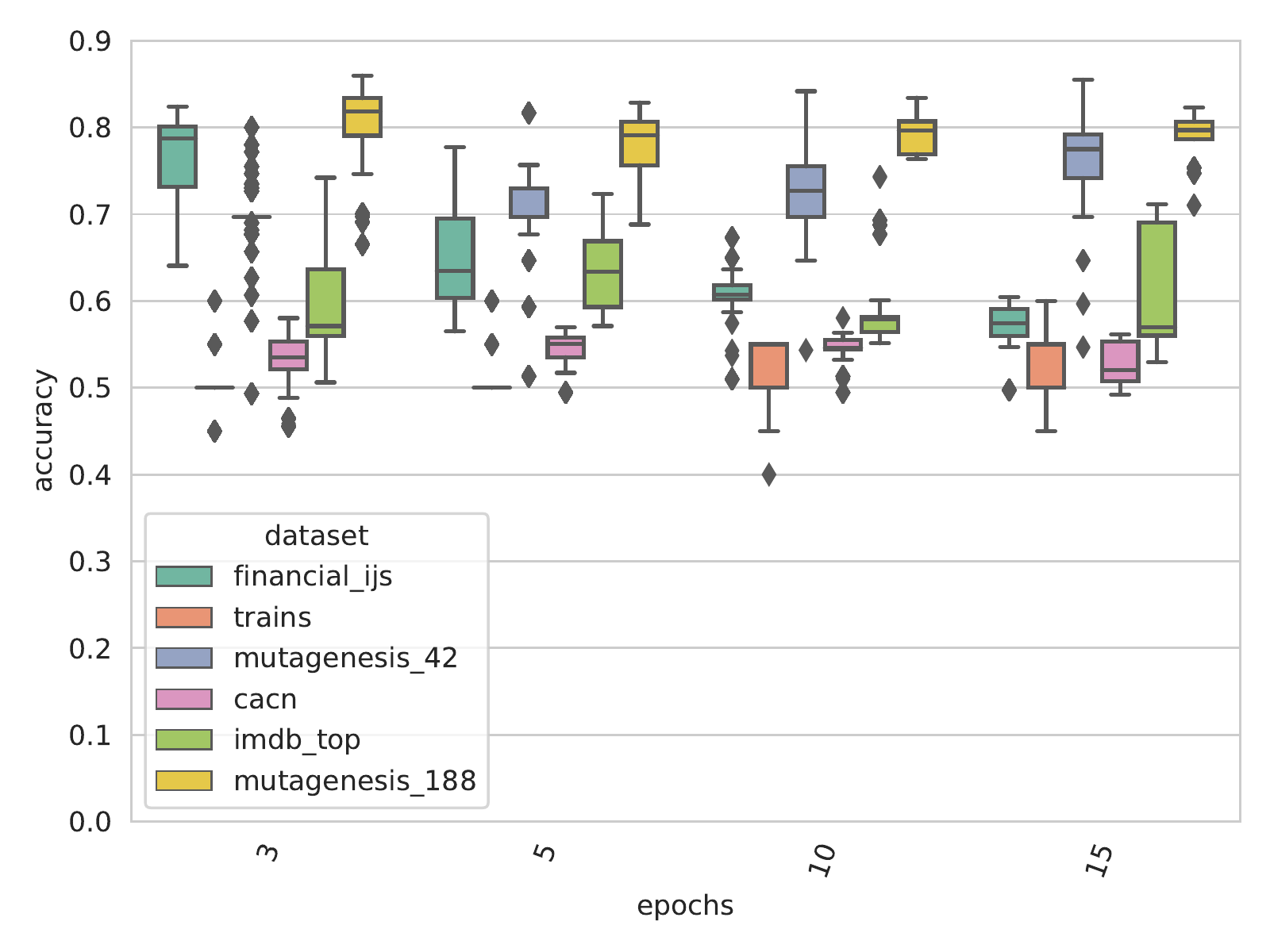}} \\
\subcaptionbox{Dependence on the learning rate.}{\includegraphics[width = 2.2in]{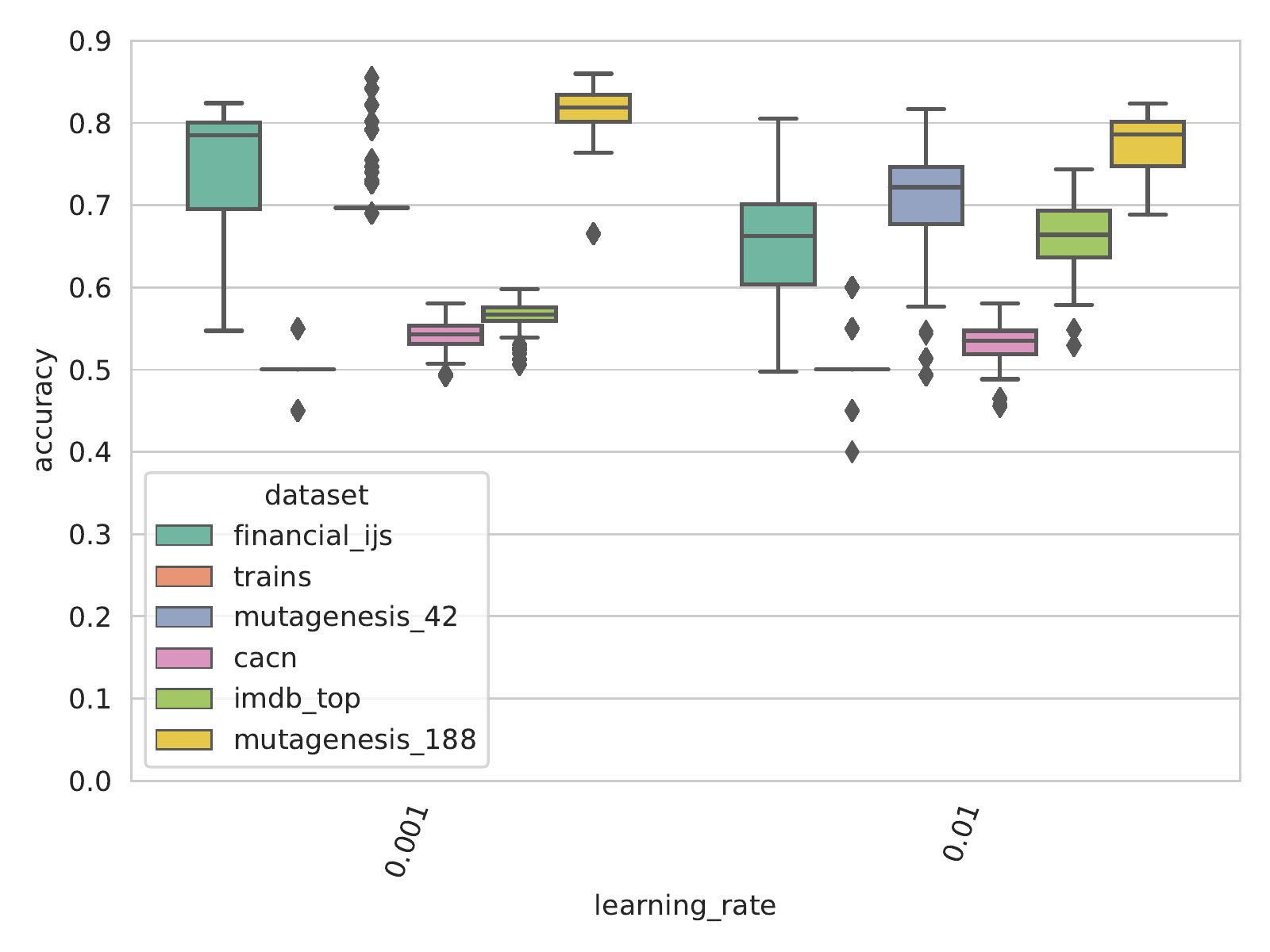}} &
\subcaptionbox{Dependence on Dropout.}{\includegraphics[width = 2.2in]{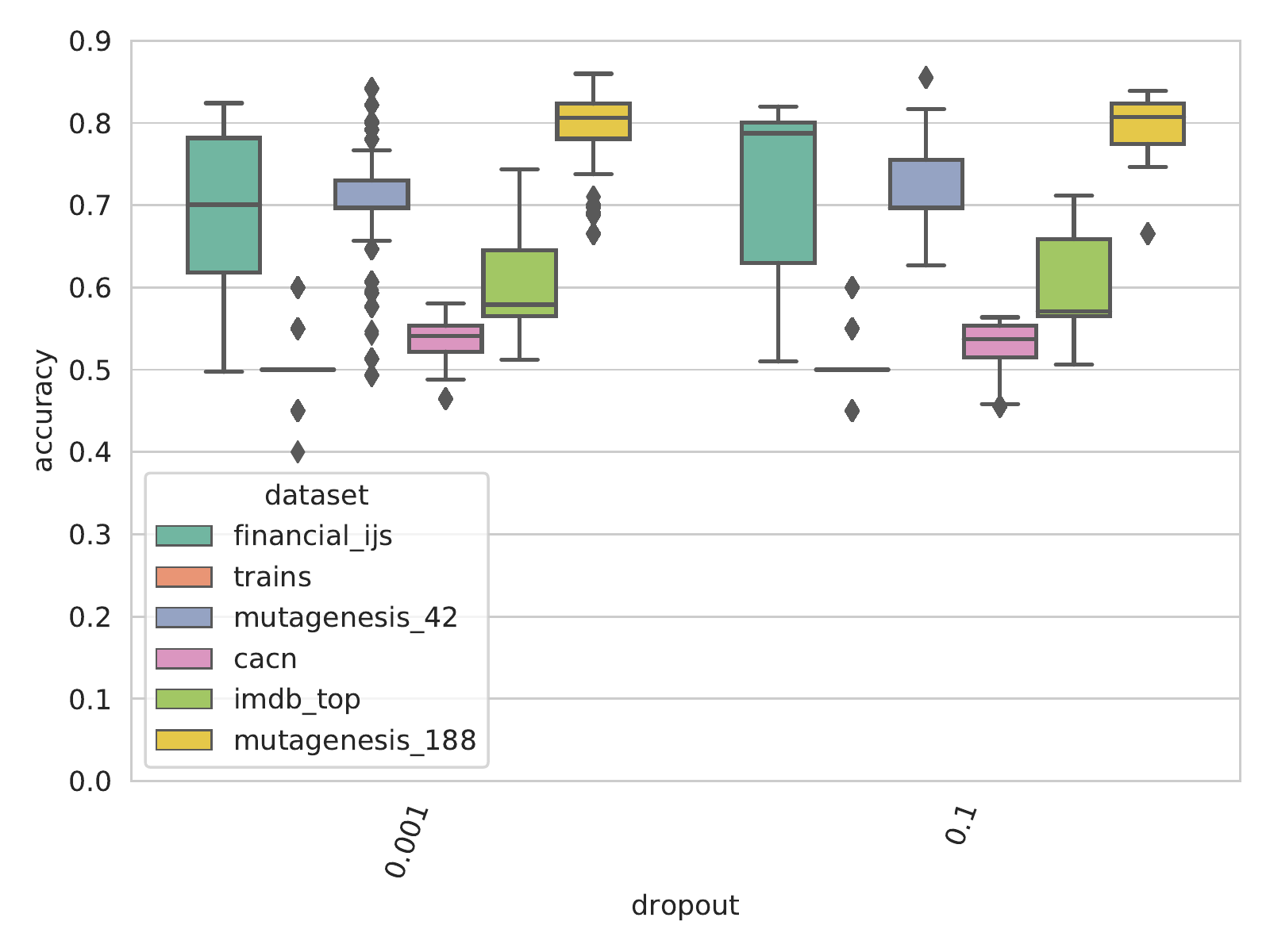}} \\
\end{tabular}
\caption{Sensitivity of PropDRM to hyperparameter settings.}
\label{abl:drm}
\end{figure}

We discuss the impact of individual hyperparameters on the performance of PropDRM. We first visualize the performance of PropDRM w.r.t. individual hyperparameters in Figure~\ref{abl:drm}.

We can observe that the relevance of individual hyperparameters varies from data set to data set. 
The learning rate, when too small, decreases the performance. In terms of embedding dimension, even smaller dimensions are sufficient for the considered data sets. This result potentially implies that the considered data sets are relatively small and contain only a small set of relevant features (when propositionalized). Thus, if the neural network detects the correct features as relevant, not many parameters are needed for a successful classification. An alternative explanation would imply that PropDRM learns hierarchical representations efficiently, albeit not optimized with their hierarchical nature in mind, which was previously demonstrated to capture hierarchical relations well \cite{nickel2017poincare}.

\section{Ablation study---PropStar}
\label{appendix:ablation-dp}

\begin{figure}[h]
\centering
\captionsetup{width=\linewidth}
\begin{tabular}{cc}
\subcaptionbox{Dependence on embedding dimensionality.}{\includegraphics[width = 2.2in]{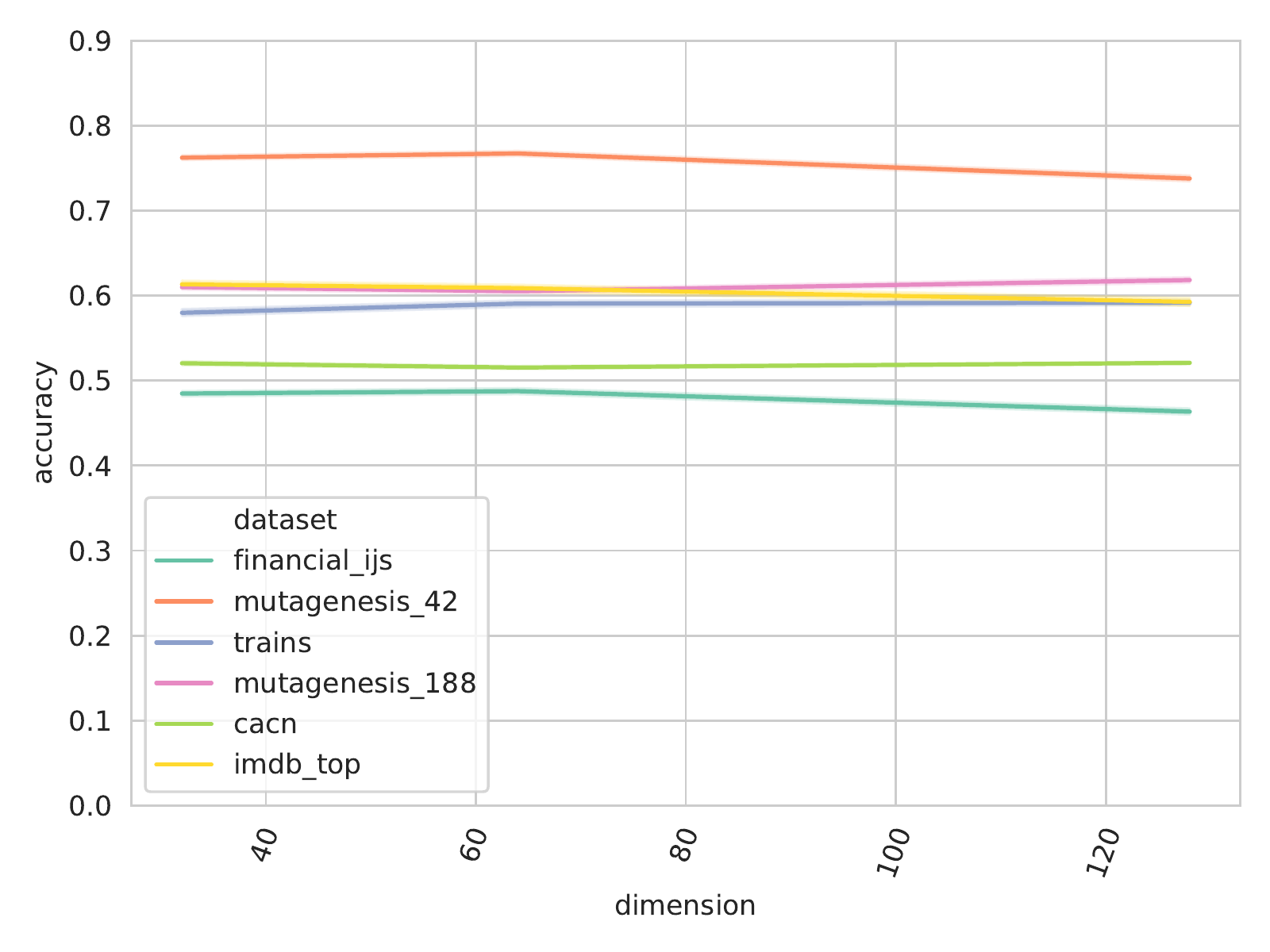}} &
\subcaptionbox{Dependence on the number of epochs.}{\includegraphics[width = 2.2in]{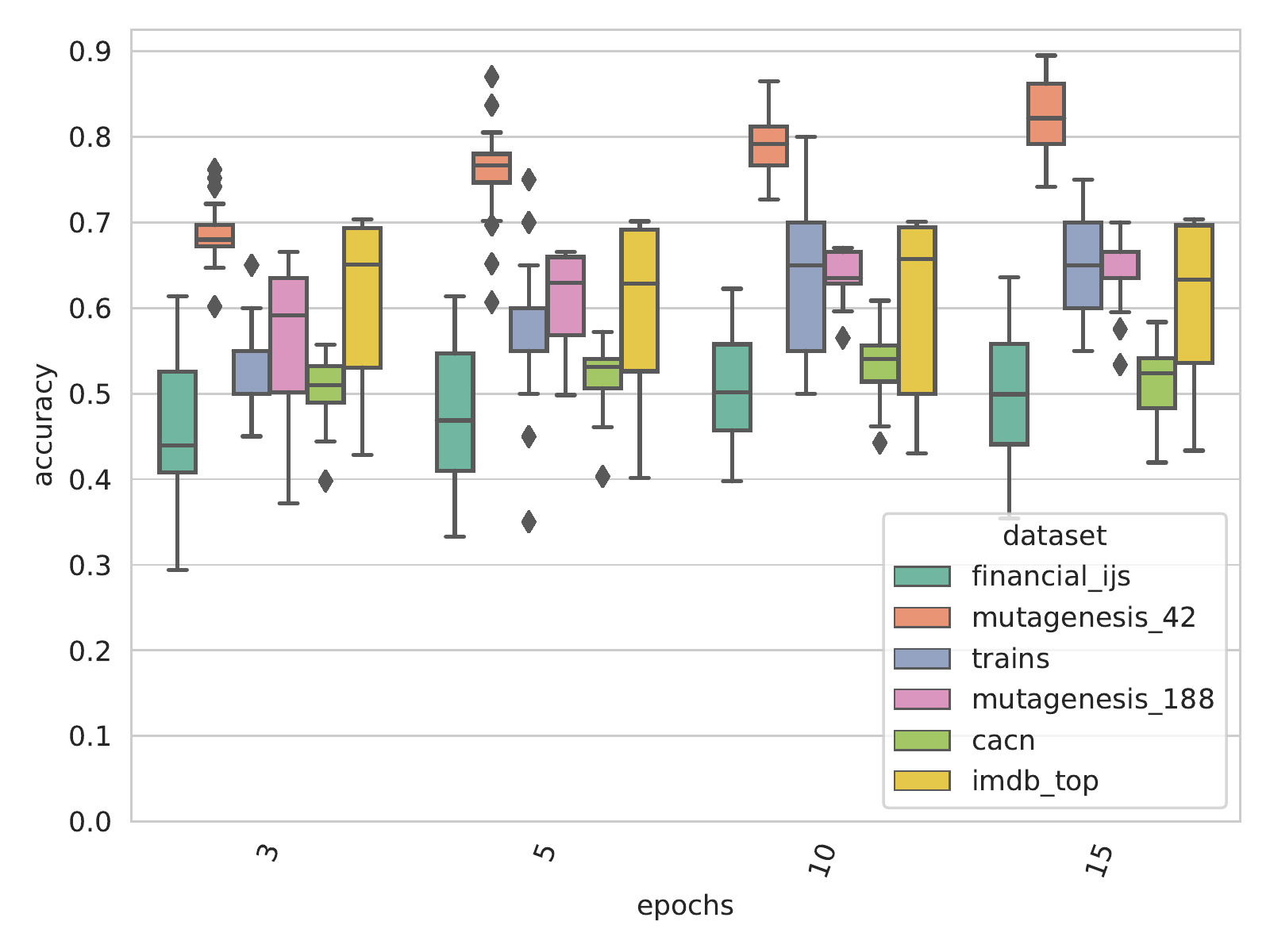}} \\
\subcaptionbox{Dependence on the learning rate.}{\includegraphics[width = 2.2in]{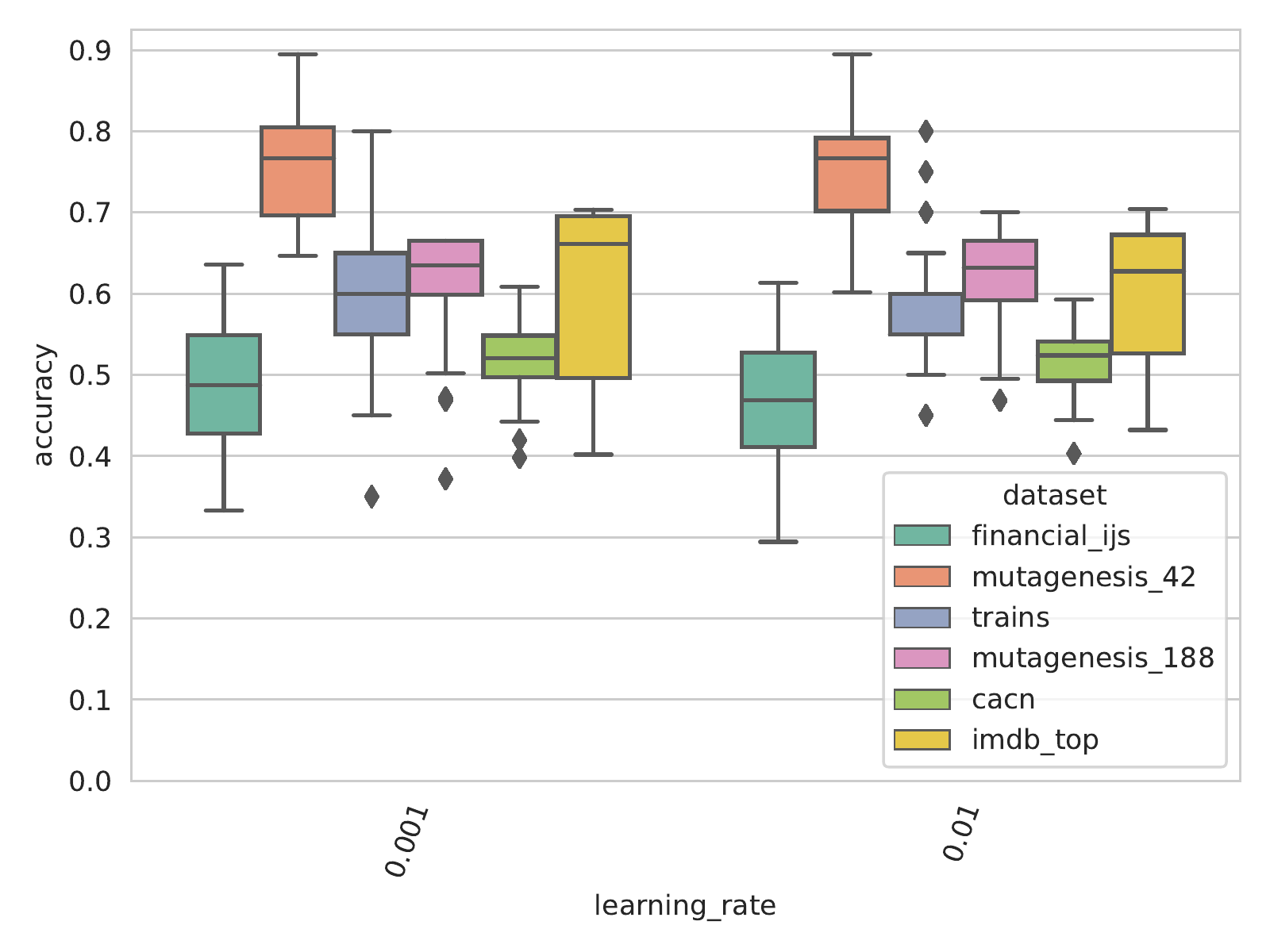}} &
\subcaptionbox{Dependence on the maximum negative sampling number.}{\includegraphics[width = 2.2in]{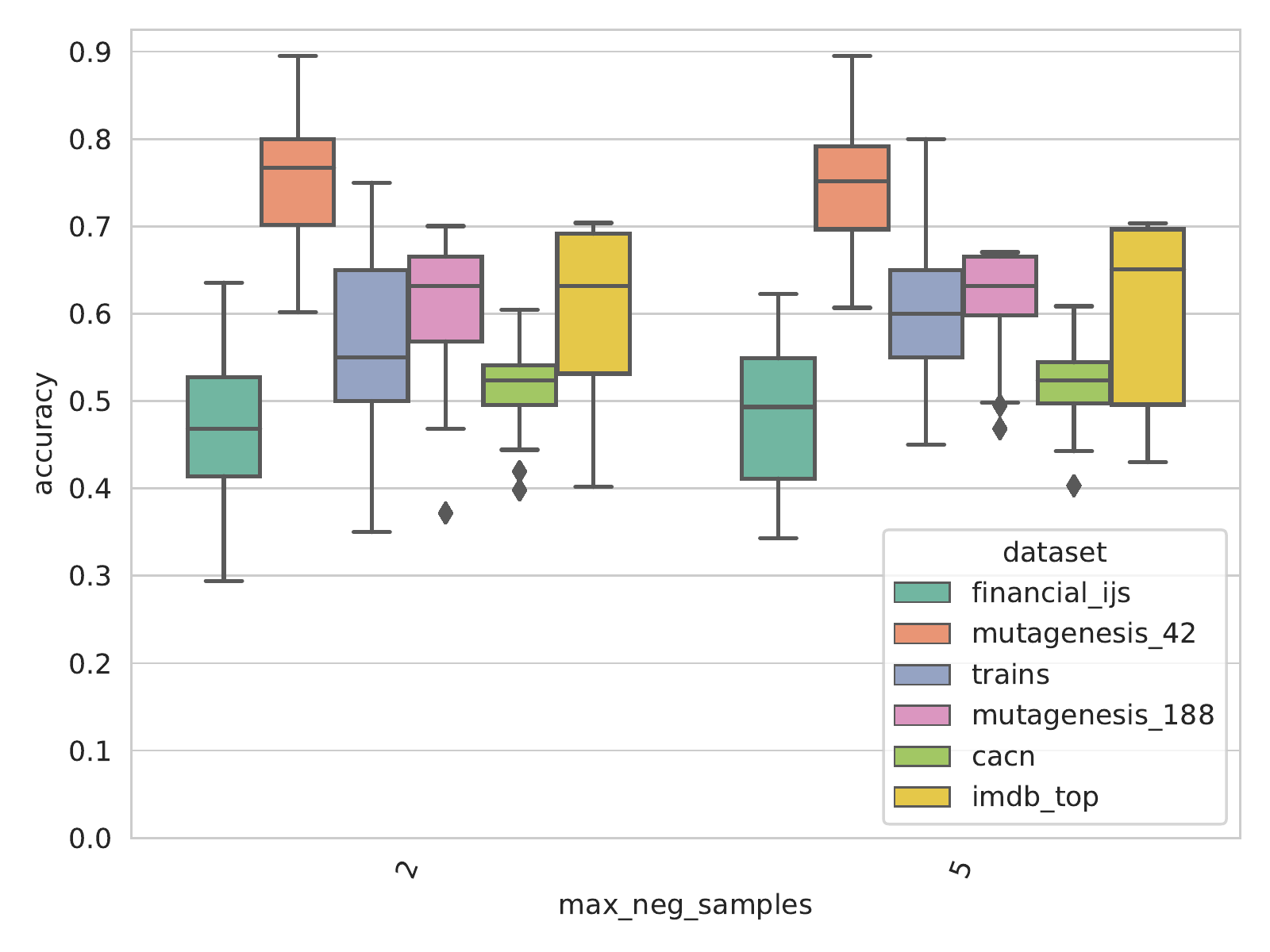}} \\
\end{tabular}
\caption{Sensitivity of PropStar to various hyperparameter settings.}
\label{abl:propembed}
\end{figure}

We first explore the behavior with respect to various hyperparameter settings and visualize them in Figure~\ref{abl:propembed}.
We can observe that the amount of negative samples (Subfigure  \ref{abl:propembed}d)) impacts the PropStar's performance the most on the mutagenesis 42 data set, overall reducing the performance, even though a handful of models (outliers marked as dots) perform well. This indicates the importance of negative sample selection.  As StarSpace does not use any sophisticated technique for sampling negative examples, the variability in performance could be notable due to this parameter.

It can be observed that a relatively small relational item embedding dimensionality is needed for successful performance. The dependence on other parameters varies from data set to data set. For example, the learning rate does not impact the larger Mutagenesis data set (Mut188) as much as it does the Trains data set. As the proposed methodology is not well adapted to such small data sets (e.g., tens of instances), large variability in performance could be linked to potential overfitting. Further, sufficient number of epochs are needed for PropStar to converge on individual data sets.

\section{Interpretability of embedding-based methods using SHAP}
\label{sec:interpretability}
The approximation power of deep neural network which are commonly used with embeddings comes at a cost of lesser interpretability. Compared to symbolic relational (or propositional) learners, one cannot understand the deep relational models' deductive process by  inspecting the model. However, \emph{post hoc} explanation methods for prediction models can be used to better understand which parts of the feature space are relevant for the neural network's individual predictions. In this work, we leverage the state-of-the-art explanation tool SHAP \cite{shap}, based on the coalitional game theory. This tool captures the importance of interactions between features with Shapley values.

When considered in a feature importance scenario, the contribution of the $i$-th instance $\tau_i$, is approximated by SHAP with the following expression:
\begin{equation*}
    \tau_i = \underbrace{\sum_{S \subseteq F \setminus \{i\}} \frac{|S|!(|F| - |S| - 1)!}{|F|!}}_{\textrm{All possible subsets}} 
    \underbrace{\bigg [   f(x_{S \cup \{i\}}) - f(x_S)   \bigg ]}_{\textrm{Difference in predictive performance}}
\end{equation*}

\noindent where $S$ is a subset of all features $F$, $f$ is the used predictive model, and $x_S$ is an instance containing only features from the subset $S$.
Shapley valufs offer insights into instance-level predictions by assigning fair credit to individual features for participation in interactions. They are commonly used to understand and debug black-box models. 

In this work, we use the SHAP kernel approximator, the recently introduced, model-agnostic method for explaining model outputs.
The used SHAP kernel explainer is considered an additive feature attribution method. Such methods are characterized as having an explanation model $g$ that is a linear function of binary variables:
\begin{equation*}
g(z') = \phi_0 + \sum_{i = 1}^{|F|}\phi_i \cdot z_i'
\end{equation*}
\noindent where $z' \in \{0,1\}^{|F|}$, $|F|$ is the number of input features and $\theta_i \in \mathbb{R}$. This class of models assign an effect $\phi_i$ to each feature, and summing the effects of all such feature attributions approximates the output $f(x)$ of the original model.
Detailed theoretical analysis of how this idea can be extended to approximation of outputs via a kernel is given in \cite{shap}. 

\begin{figure}[h]
\centering
\captionsetup{width=\linewidth}
\begin{tabular}{cc}
\subcaptionbox{Mean of SHAP explanations over the instances for PropDRM.}{\includegraphics[width = 2.2in, height = 3in]{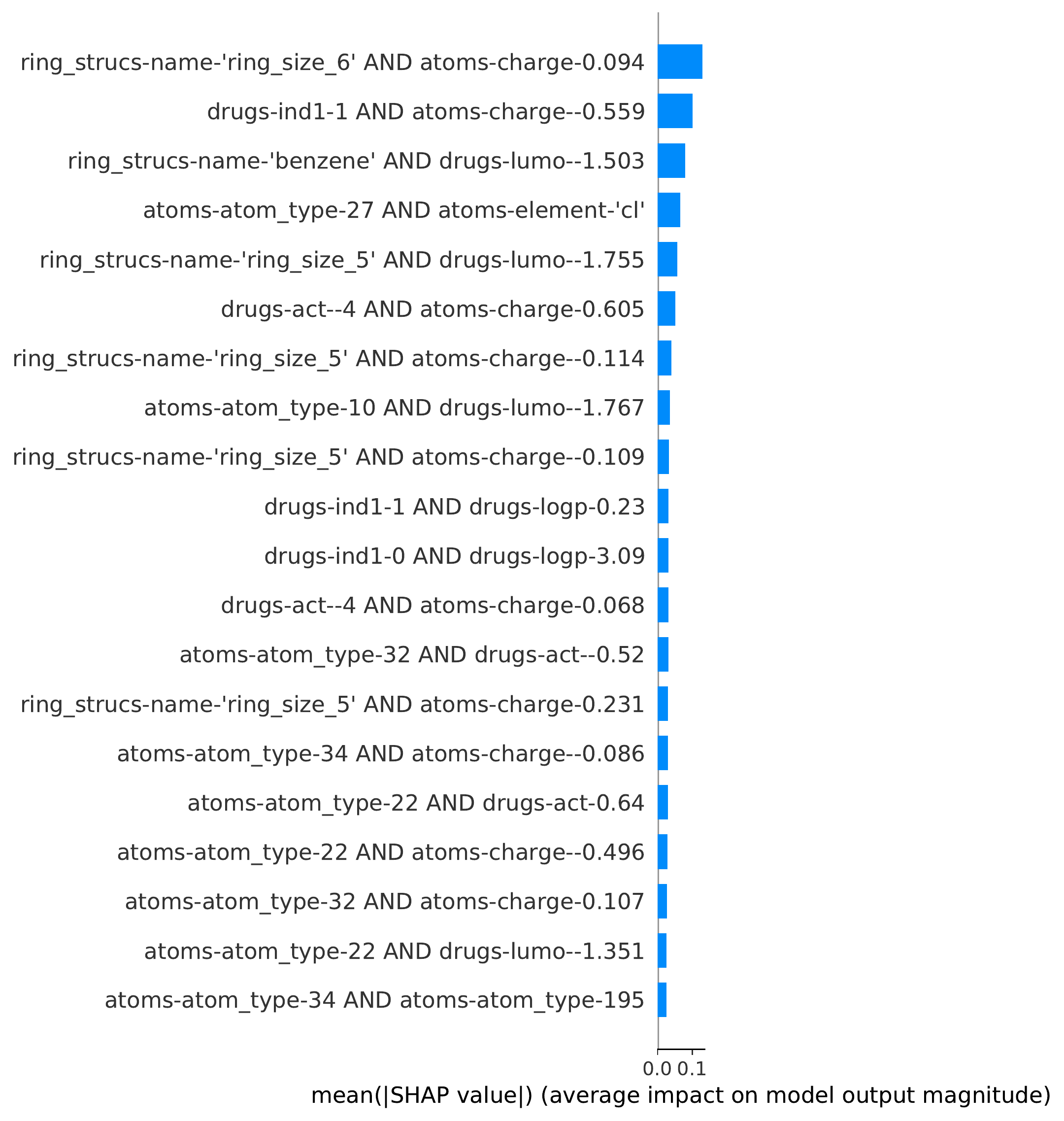}} & 
\subcaptionbox{Mean of SHAP explanations over the instances for PropStar.}{\includegraphics[width = 2.2in, height = 3in]{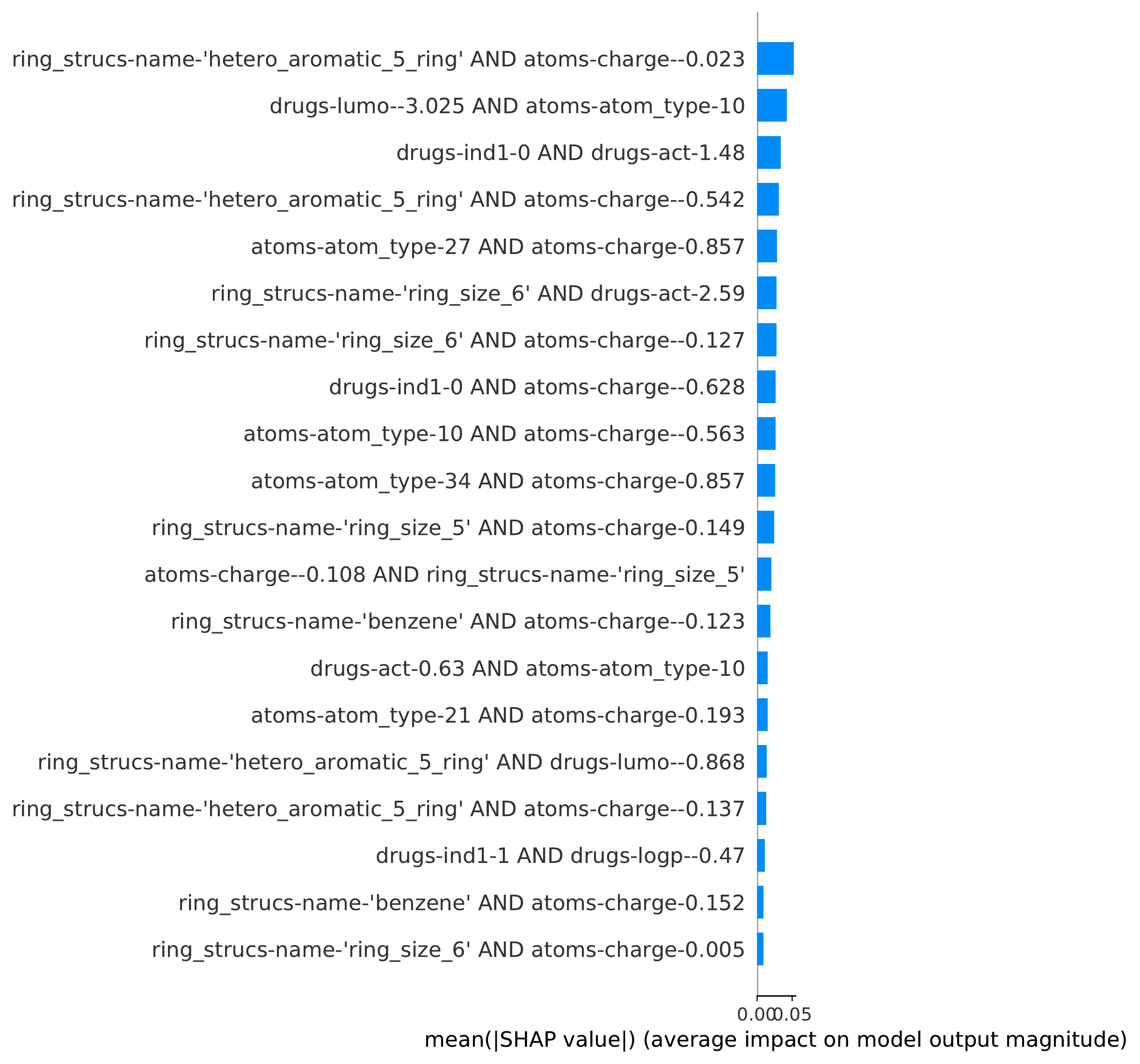}}
 \\
\end{tabular}
\caption{SHAP Kernel explanations of the two developed approaches.}
\label{figure:shap-pemb}
\end{figure}

As an example demonstrating the explainability of the two paradigms, we visualize the Shapley values as explanations of learned classifiers for Mutagenesis 188 problem in Figure~\ref{figure:shap-pemb}. Explanations indicate parts of the feature space that have the largest impact on the model's output. 
Even though the considered SHAP kernel explainer is known to be a computationally  expensive variant of SHAP (it is also the most general one), explanations were obtained in the order of minutes, indicating the potential of this methodology for explanations of predictors in larger relational databases.

\end{document}